\documentclass{llncs}
\usepackage{times}

\usepackage{latexsym}

\usepackage{graphicx}
\usepackage{wrapfig}

\usepackage{xspace}

\begin{document}

\title{Breaking Symmetry with Different Orderings}

\author{Nina Narodytska \and Toby Walsh}
\institute{NICTA and UNSW,\\
Sydney, Australia\\
email: \{nina.narodytska,toby.walsh\}@nicta.com.au}

\maketitle
\sloppy
\begin{abstract}
We can break symmetry
by eliminating
solutions within each symmetry class.
For instance, the Lex-Leader method
eliminates all but the
smallest solution in the lexicographical ordering.
Unfortunately, %as symmetry groups can be large,
the Lex-Leader method is intractable in general.
We prove that, under modest assumptions,
we cannot reduce the worst case complexity of breaking symmetry by
using other orderings on solutions.
We also prove that a common type of symmetry, where rows and columns
in a matrix of decision variables are interchangeable,
is intractable to break when we use two promising alternatives
to the lexicographical ordering:
the Gray code ordering (which uses a different ordering
on solutions), and the Snake-Lex
ordering (which is a variant of the lexicographical ordering
that re-orders the variables).
Nevertheless, we
show experimentally that using other
orderings like the Gray
code to break symmetry can be beneficial in practice
as
%They can pick out different solutions in each symmetry class
%that
they may better align with
the objective function and branching
heuristic.
\end{abstract}

\newcommand{\DLex}{\mbox{\sc DoubleLex}}
\newcommand{\snakelex}{\mbox{\sc SnakeLex}}

\newcommand{\set}{\mathcal}
\newcommand{\myset}[1]{\ensuremath{\mathcal #1}}

\renewcommand{\theenumii}{\alph{enumii}}
\renewcommand{\theenumiii}{\roman{enumiii}}
\newcommand{\figref}[1]{Figure \ref{#1}}
\newcommand{\tref}[1]{Table \ref{#1}}
\newcommand{\myldots}{\ldots}

\newtheorem{myproblem}{Problem}
\newtheorem{mydefinition}{Definition}
\newtheorem{mytheorem}{Proposition}
\newtheorem{mylemma}{Lemma}
\newtheorem{myexample}{Running Example}{\bf}{\it}
\newtheorem{mytheorem1}{Theorem}
\newcommand{\myproof}{\noindent {\bf Proof:\ \ }}
\newcommand{\myqed}{\mbox{$\Box$}}
\newcommand{\myend}{\mbox{$\clubsuit$}}

\newcommand{\mymod}{\mbox{\rm mod}}
\newcommand{\mymin}{\mbox{\rm min}}
\newcommand{\mymax}{\mbox{\rm max}}
\newcommand{\range}{\mbox{\sc Range}}
\newcommand{\roots}{\mbox{\sc Roots}}
\newcommand{\myiff}{\mbox{\rm iff}}
\newcommand{\alldifferent}{\mbox{\sc AllDifferent}}
\newcommand{\permutation}{\mbox{\sc Permutation}}
\newcommand{\disjoint}{\mbox{\sc Disjoint}}
\newcommand{\cardpath}{\mbox{\sc CardPath}}
\newcommand{\CARDPATH}{\mbox{\sc CardPath}}
\newcommand{\common}{\mbox{\sc Common}}
\newcommand{\uses}{\mbox{\sc Uses}}
\newcommand{\lex}{\mbox{\sc Lex}}
\newcommand{\usedby}{\mbox{\sc UsedBy}}
\newcommand{\nvalue}{\mbox{\sc NValue}}
\newcommand{\slide}{\mbox{\sc CardPath}}
\newcommand{\sliden}{\mbox{\sc AllPath}}
\newcommand{\SLIDE}{\mbox{\sc CardPath}}
\newcommand{\circularslide}{\mbox{\sc CardPath}_{\rm O}}
\newcommand{\among}{\mbox{\sc Among}}
\newcommand{\mysum}{\mbox{\sc MySum}}
\newcommand{\amongseq}{\mbox{\sc AmongSeq}}
\newcommand{\atmost}{\mbox{\sc AtMost}}
\newcommand{\atleast}{\mbox{\sc AtLeast}}
\newcommand{\element}{\mbox{\sc Element}}
\newcommand{\gcc}{\mbox{\sc Gcc}}
\newcommand{\gsc}{\mbox{\sc Gsc}}
\newcommand{\contiguity}{\mbox{\sc Contiguity}}
\newcommand{\PRECEDENCE}{\mbox{\sc Precedence}}
\newcommand{\assignnvalues}{\mbox{\sc Assign\&NValues}}
\newcommand{\linksettobooleans}{\mbox{\sc LinkSet2Booleans}}
\newcommand{\domain}{\mbox{\sc Domain}}
\newcommand{\symalldiff}{\mbox{\sc SymAllDiff}}
\newcommand{\alldiff}{\mbox{\sc AllDiff}}

\newcommand{\slidingsum}{\mbox{\sc SlidingSum}}
\newcommand{\MaxIndex}{\mbox{\sc MaxIndex}}
\newcommand{\REGULAR}{\mbox{\sc Regular}}
\newcommand{\regular}{\mbox{\sc Regular}}
\newcommand{\precedence}{\mbox{\sc Precedence}}
\newcommand{\STRETCH}{\mbox{\sc Stretch}}
\newcommand{\SLIDEOR}{\mbox{\sc SlideOr}}
\newcommand{\NAE}{\mbox{\sc NotAllEqual}}
\newcommand{\mytheta}{\mbox{$\theta_1$}}
\newcommand{\mysigma}{\mbox{$\sigma_2$}}
\newcommand{\mysigmatwo}{\mbox{$\sigma_1$}}

\newcommand{\todo}[1]{{\tt (... #1 ...)}}
\newcommand{\myOmit}[1]{}
\newcommand{\nina}[1]{#1}
\newcommand{\ninacpp}[1]{#1}
\newcommand{\ninacp}[1]{#1}

\newcommand{\dpsb}{DPSB}

\sloppy
\section{Introduction}

Symmetry occurs in many combinatorial
problems.
For example, when coloring a graph, we can permute
the colors in any proper coloring.
Symmetry can also be introduced by modelling
decisions (e.g. using a set of finite domain variables
to model a set of objects will introduce the symmetries that
permute these variables).
A common method to deal with symmetry is to
add constraints which eliminate symmetric
solutions
%(e.g. \cite{puget:Sym}).
%\cite{puget:Sym,ssat2001,ffhkmpwcp2002,llconstraints06,wecai2006,wcp07,llwycp07,knwcp10}.
%\cite{puget:Sym,ssat2001,ffhkmpwcp2002,llconstraints06,wecai2006}.
%\cite{puget:Sym,ssat2001,llconstraints06,wecai2006,wcp06,llwycp07}.
(e.g.
\cite{puget:Sym,clgrkr96,ssat2001,asmijcai2003,llcp2004,pijcai2005,llconstraints06,wecai2006,wcp06,llwycp07,waaai2008,knwcp10,handbookcp}).
%\cite{puget:Sym}.
Unfortunately, breaking symmetry by adding constraints
to eliminate symmetric solutions
is intractable in general \cite{clgrkr96}.
More specifically, deciding if an assignment is the
smallest in its symmetry class
for a matrix with row and column
symmetries is NP-hard, supposing
rows are appended together
and compared lexicographically.
There is, however, nothing
special about appending rows together or
comparing solutions lexicographically. We
could use {\em any} total ordering
over assignments.
For example, we could break symmetry
with the Gray code ordering. That is,
we add constraints that eliminate symmetric
solutions within each symmetry
class that are not smallest in the Gray code ordering.
This is a total ordering over
assignments used in error correcting codes.
Such an ordering may pick out different
solutions in each symmetry class,
reducing the conflict between
symmetry breaking, problem constraints, objective
function and the branching heuristic. %We consider
%ahere the binary Gray code but note that it can be generalized
%to deal with non-binary domains.
%
The Gray code ordering has some properties that
may be useful for symmetry breaking. In particular,
neighbouring assignments in the ordering only differ
at one position, and flipping one bit reverses
the ordering of the subsequent bits.
%It is therefore not clear if breaking
%symmetry with the Gray code ordering is also intractable.
%We shall show that it is.

As a second example, we can break row and
column symmetry
%in a matrix model
with the Snake-Lex ordering \cite{snakelex}.
This orders assignments %within a symmetry class
by lexicographically comparing
vectors constructed by appending the variables in the matrix
in a ``snake like'' manner. The
\ninacpp{first row is appended to the reverse
of the second row, and this is then
appended to the third row, and then the
reverse of the fourth row and so on.}
%Again, this picks out different solutions
%in each symmetry class.
As a third example, %it is often effective to
we can break row and column symmetry
%in a matrix model
by ordering the rows
lexicographically and the columns with
a multiset ordering \cite{fhkmwijcai2003}.
This is incomparable to the
Lex-Leader method.

%Does using different
%orderings like the Gray code and Snake-Lex
%ordering change the computational complexity
%of breaking symmetry?
We will argue theoretically that breaking
symmetry with a different ordering over
assignments cannot improve the worst case complexity.
However, we also show that other orderings
can be useful in practice as they pick out different solutions
in each symmetry class.
Our argument has two parts.
We first argue that, under modest assumptions which
are satisfied by the Gray code and Snake-Lex orderings,
we cannot reduce the computational complexity from that
of breaking symmetry with the lexicographical ordering
which considers variables in a matrix row-wise.
We then prove that for the particular case of row and column
symmetries, breaking symmetry with the Gray code or Snake-Lex
ordering is intractable (as it was with the
lexicographical ordering).
Many dynamic methods for dealing with symmetry
are equivalent to posting symmetry breaking
constraints ``on the fly''
(e.g.
\cite{bscade92,bsjar94,backofen:Sym,sbds,fahle1,sellmann2,dynamiclex,kwecai10,nwcpaior13}).

%%\cite{backofen:Sym,sbds,fahle1,sellmann2,dynamiclex}).
%\cite{backofen:Sym,sbds,dynamiclex}.
%(e.g. \cite{backofen:Sym,sbds}).
Hence, our results have implications for such
dynamic methods too.

\section{Background}

A symmetry of a set of constraints $S$ is a bijection $\sigma$
on complete assignments that maps solutions of $S$ onto
other solutions of $S$. Many of our results apply to the
more restrictive %(and perhaps usual)
definition of symmetry which
considers just those bijections which
map individual variable-value pairs
\cite{cjjpsconstraints06}.
However, this more general definition captures
also conditional symmetries \cite{gklmmscp05}.
In addition, a few of our results require this
more general definition. In particular, Theorem 3
only holds for this more general definition\footnote{We thank
an anonymous reviewer for pointing this out.}.
%We lift $\sigma$ from variable-value pairs
%to complete assignments (and hence solutions) in the
%natural way: $\sigma(A) = \{ \sigma(X=v) \ | \ X=v \in A\}$.
%Thus, a bijection $\sigma$ on variable-value pairs
%is a {\em symmetry} of a set of constraints
%$S$ iff given any solution $A$ of $S$, $\sigma(A)$ is also a solution
%of $S$.
%A {\em variable symmetry} is a bijection that just
%acts on the variable indices, whilst a {\em value symmetry}
%is a bijection that just acts on the values.
The set of symmetries form a group under composition.
Given a symmetry group $\Sigma$, a subset $\Pi$
generates $\Sigma$ iff any $\sigma \in \Sigma$ is a composition
of elements from $\Pi$.
A symmetry group $\Sigma$ partitions the solutions
into symmetry classes (or orbits). We write $[A]_{\Sigma}$ for
the symmetry class of solutions symmetric to
the solution $A$. Where $\Sigma$ is clear from the
context, we write $[A]$.
%Note that symmetry classes
%are equivalence classes.
A set of symmetry
breaking constraints is {\em sound} iff it
leaves at least one solution in each symmetry
class, and {\em complete}
iff it leaves at most one solution in each symmetry
class.

We will study what happens to symmetries when problems
are reformulated onto equivalent problems. For example, we
might consider the Boolean form of a problem in which
$X_i=j$ maps onto $Z_{ij}=1$.
Two sets of constraints, $S$ and $T$ over
possibly different variables
are {\em equivalent} iff there is a bijection %$\pi$
between their solutions. % of $S$ and of $T$.
Suppose $U_i$ and $V_i$ for $i \in [1,k]$ are partitions of
the sets $U$ and $V$ into $k$ subsets.
Then the two partitions are {\em isomorphic}
iff there are bijections $\pi: U \mapsto V$
and $\tau:[1,k]\mapsto [1,k]$ such
that $\pi(U_i)=V_{\tau(i)}$ for $i \in [1,k]$
where $\pi(U_i) = \{ \pi(u) \ | \ u \in U_i \}$.
Two symmetry groups $\Sigma$ and $\Pi$
of constraints $S$ and $T$ respectively are {\em isomorphic} iff
$S$ and $T$ are equivalent, and
their symmetry classes of solutions are isomorphic.
%When two symmetry groups are isomorphic,
%the number and sizes of their symmetry
%soclasses are identical.

\section{Using Other Orderings}

The Lex-Leader method \cite{clgrkr96}
picks out the lexicographically smallest solution in each
symmetry class. For every symmetry $\sigma$,
it posts a lexicographical ordering constraint:
$ \langle X_1, \ldots, X_n \rangle \leq_{\rm lex}
\sigma (\langle X_1, \ldots, X_n \rangle)
$
where $X_1$ to $X_n$ is some ordering on the
variables in the problem.
Many static symmetry breaking constraints
can be derived from such Lex-Leader constraints.
For example, {\sc DoubleLex} constraints
to break row and column symmetry can be
derived from them \cite{lex2001}.
As a second example,
{\sc Precedence} constraints
to break the symmetry due to interchangeable
values can also be
derived from them  \cite{llcp2004,wecai2006}.
%\cite{llcp2004}.
Efficient algorithms exist to propagate
such lexicographical % ordering
constraints %\cite{fhkmwcp2002}.
(e.g. \cite{fhkmwcp2002,fhkmwaij06,knwercim09}).

We could, however, break symmetry by using
another ordering on assignments like the Gray code
ordering. \ninacpp{We define the Gray code ordering on Boolean variables.}
 For each symmetry $\sigma$,
we could post an ordering constraint:
$$ \langle X_1, \ldots, X_n \rangle \leq_{\rm Gray}
\sigma (\langle X_1, \ldots, X_n \rangle)
$$
Where the $k$-bit Gray code ordering 
is defined recursively as follows: 0 is before 1,
and to construct the $k+1$-bit ordering,
we append 0 to the front of the $k$-bit ordering,
and concatenate it with the reversed $k$-bit ordering with 1 appended
to the front.
For instance, the
%\myOmit{
4-bit Gray code orders assignments as follows:
\begin{eqnarray*}
&0000,
0001,
0011,
0010,
0110,
0111,
0101,
0100,& \\
&\ \ \ \ \ \ \ \ 1100,
1101,
1111,
1110,
1010,
1011,
1001,
1000
&
\end{eqnarray*}
%}
\myOmit{
3-bit Gray code orders assignments as follows:
\begin{eqnarray*}
&000, \
001, \
011, \
010, \
110, \
111, \
101, \
100
&
\end{eqnarray*}}
The Gray code ordering is well founded. Hence,
every set of complete assignments will have a smallest
member under this ordering. This is the unique
complete assignment in each symmetry class selected
by posting such Gray code ordering constraints.
Thus breaking symmetry with Gray code ordering
constraints is
sound and complete.

\begin{mytheorem}
Breaking symmetry with Gray code ordering
constraints is sound and complete.
\end{mytheorem}
%\myproof
%To show soundness and completeness, we note that
%The Gray code ordering is well founded. Hence,
%every set of complete assignments will have a smallest
%member under this ordering. This is the unique
%complete assignment in each symmetry class selected
%by posting such Gray code ordering constraints.
%\myqed

In Section 6, we propose a propagator for
the Gray code ordering constraint. %Unlike
%the Snake-Lex ordering, % \cite{snakelex},
We cannot enforce the Gray code ordering
by ordering variables and values,
and using a lexicographical ordering constraint.
For example, we cannot map the 2-bit
Gray code onto the lexicographical ordering
by simply re-ordering variables and values.
To put it another way, no reversal and/or inversion
of the bits in the 2-bit Gray code will map it onto
the lexicographical ordering.
The 2-bit Gray code orders 00, 01, 11 and then 10.
We can invert the first bit to give:
10, 11, 01 and then 00.
Or we can invert the second bit to give:
01, 00, 10, and then 11.
Or we can invert both bits to give:
11, 10, 00, and then 01.
We can also reverse the bits
to give:
00, 10, 11, and then 01.
And we can then invert one or both bits to
give:
10, 00, 01, and then 10; or
01, 11, 10, and then 00; or
11, 01, 00, and then 10.
Note that none of these re-orderings and
inversions is the 2-bit lexicographical
ordering: 00, 01, 10, and then 11.

\myOmit{
In Section 6, we propose a propagator for
a Gray code ordering constraint as, unlike
other new orderings proposed in the past
like Snake-Lex \cite{snakelex},
we cannot enforce the Gray code ordering
simply by reordering variables and values,
and using the lexicographical ordering.

\begin{mytheorem}
We cannot map the Gray code
ordering onto the lexicographical ordering
by reordering variables and values.
\end{mytheorem}
\myproof
Consider the 2 bit lexicographical ordering:
$00,01,10,11$.
We can reorder the first value to get $10,11,01,01$,
or the second to get $01,00,11,10$,
or both values to get $11,10,01,00$.
Alternatively, we can swap variables to get
$00,10,01,11$,
or swap variables and reorder the first value to get $10,00,11,01$,
or swap variables and reorder the second value to get $01,11,00,10$,
or swap variables and reorder both values to get $11,01,10,00$.
None of these re-orderings %of variables and values
%in the lexicographical ordering
gives the 2-bit Gray code ordering: $00,01,11,10$.
\myqed
}

\section{Complexity of Symmetry Breaking}

\myOmit{
We first argue that other
orderings besides the lexicographical ordering
can increase the computational
complexity of symmetry breaking. Our argument has two
parts. First, even with
a single symmetry, we can introduce
%computational
complexity through
the complexity of deciding the ordering. Second,
even if the ordering is polynomial to decide,
we can introduce %computational
complexity
by having many symmetries to break.

\begin{mylemma}
There exists a total ordering $\preceq$ on assignments,
and a class of problems $P$ that is polynomial to decide
and has a single symmetry $\sigma$ such
that finding a solution of $\{ S \cup
\{ \langle X_1, \ldots, X_n \rangle \preceq
\sigma (\langle X_1, \ldots, X_n \rangle) \} \ | \ S \in P\}$
is NP-hard,
whilst finding a solution of $\{ S \cup
\{ \langle X_1, \ldots, X_n \rangle \leq_{\rm lex}
\sigma (\langle X_1, \ldots, X_n \rangle) \} \ | \ S \in P\}$
is polynomial.
\end{mylemma}
\myproof
Reduction from 1-in-3-SAT on $m$ positive
clauses.
% In 1-in-3-SAT, we wish to decide
%if 3-cnf formula can be satisfied by a truth
%assignment that sets exactly 1 out of the 3 literals
%in each clause to true.
Let $n=3m+1$,
and $S$ be a set of
unary constraints that ensure $X_{3p-2}=i$, $X_{3p-1}=j$, $X_{3p}=k$
where $p \in [1,m]$ and the $p$th clause is $i \vee j \vee k$.
We also have $X_{3m+1} \in \{0,1\}$.
Consider the symmetry $\sigma$ that
interchanges the values $0$ and $1$ for
$X_{3m+1}$.
We define an ordering $\prec$ as follows.
$\langle X_1, \ldots, X_{3m+1} \rangle
\prec \langle Y_1, \ldots, Y_{3m+1} \rangle$
iff one of 3 conditions holds:
%\begin{enumerate}
%\item
1) $\langle X_1, \ldots, X_{3m} \rangle
<_{\rm lex} \langle Y_1, \ldots, Y_{3m} \rangle$,
or
%\item
2) $\langle X_1, \ldots, X_{3m} \rangle
= \langle Y_1, \ldots, Y_{3m} \rangle$,
$X_{3m+1}=0$, $Y_{3m+1}=1$,
and the 1-in-3-SAT problem defined
by $X_1$ to $X_{3m}$ is satisfiable,
or
%\item
3)
$\langle X_1, \ldots, X_{3m} \rangle
= \langle Y_1, \ldots, Y_{3m} \rangle$,
$X_{3m+1}=1$, $Y_{3m+1}=0$,
and the 1 in 3-SAT problem defined
by $X_1$ to $X_{3m}$ is unsatisfiable.
%\end{enumerate}
%Finally, we define $\preceq$ by
%$\langle X_1, \ldots, X_{3m+1} \rangle
%\preceq \langle Y_1, \ldots, Y_{3m+1} \rangle$
%iff
%$\langle X_1, \ldots, X_{3m+1} \rangle
%\prec \langle Y_1, \ldots, Y_{3m+1} \rangle$
%or
%$\langle X_1, \ldots, X_{3m+1} \rangle
%= \langle Y_1, \ldots, Y_{3m+1} \rangle$.
We define $\vec{X} \preceq \vec{Y}$ iff
$\vec{X} \prec \vec{Y}$ or $\vec{X}=\vec{Y}$.
The only solution of
$S \cup
\{ \langle X_1, \ldots, X_{3m+1} \rangle \preceq
\langle \sigma(X_1), \ldots, \sigma(X_{3m+1}) \rangle \}$
has $X_{3m+1}=0$ if the corresponding
1-in-3-SAT problem is satisfiable
and $X_{3m+1}=1$ otherwise. Hence,
finding the solution of these constraints
is NP-hard. By comparison,
finding a solution of $S \cup
\{ \langle X_1, \ldots, X_{3m+1} \rangle \leq_{\rm lex}
\langle \sigma(X_1), \ldots, \sigma(X_{3m+1}) \rangle \}$
is
polynomial as $X_n = X_{3m+1}=0$ is always a solution.
\myqed

This proof used an
ordering that is
intractable to decide.
%More precisely, deciding if
%$\langle X_1, \ldots, X_{n} \rangle
%\preceq \langle Y_1, \ldots, Y_{n} \rangle$ was
%NP-hard.
If we insist that the ordering
is polynomial to decide,
then breaking a single symmetry will also be polynomial.
Indeed, breaking even a polynomial number of
symmetries must also be polynomial in this case.
On the other hand, with an exponential number of symmetries,
moving away from the lexicographical
ordering can sometimes increase
the computational complexity of finding a solution.

\begin{mylemma}
There exists
a total ordering $\preceq$ on assignments where
deciding $\preceq$ is polynomial, and
a class of problems $P$ that is also polynomial to decide,
and for each $S \in P$, a symmetry group $\Sigma$ of $S$
such that finding a solution of
$\{ S \cup
\{ \langle X_1, \ldots, X_n \rangle \preceq
\sigma (\langle X_1, \ldots, X_n \rangle) \ | \ \sigma \in \Sigma \} \ | \ S \in P\}$
is NP-hard,
but finding a solution of $\{ S \cup
\{ \langle X_1, \ldots, X_n \rangle \leq_{\rm lex}
\sigma (\langle X_1, \ldots, X_n \rangle) \ | \ \sigma \in \Sigma \} \ | \ S \in P\}$
is polynomial.
\end{mylemma}
\myproof
Reduction from SAT.
We let $\preceq$ be the reverse lexicographical
ordering, $\geq_{\rm lex}$. %This is
%polynomial to decide.
Consider any %SAT
formula $\varphi$ %on 0/1 variables
over $X_1$ to $X_n$.
Let $S$ be $\varphi \vee (X_1=\ldots =X_n=0)$,
and the symmetry group $\Sigma$ be such that
all solutions of $S$ are in the same symmetry class.
Consider any solution of
$S \cup
\{ \langle X_1, \ldots, X_n \rangle \preceq
\sigma(\langle X_1, \ldots, X_n \rangle) \ | \ \sigma
\in \Sigma \}$.
There are three cases. In the first,
the solution is $X_1=\ldots =X_n=0$ and this is not
a solution of $\varphi$. Then $\varphi$ is
unsatisfiable.
In the second case,
the solution is $X_1=\ldots =X_n=0$ and this is
the only solution of $\varphi$. Then $\varphi$ is
satisfiable.
In the third case,
the solution is lexicographical larger
than $X_1=\ldots =X_n=0$. Then $\varphi$ is
again satisfiable. Hence finding a solution
of $S \cup
\{ \langle X_1, \ldots, X_n \rangle \preceq
\sigma (\langle X_1, \ldots, X_n \rangle) \ | \ \sigma
\in \Sigma \}$
decides the satisfiability of $\varphi$.
%Thus
%finding a solution is NP-hard.
By comparison,
finding a solution of $S \cup
\{ \langle X_1, \ldots, X_n \rangle \leq_{\rm lex}
\sigma(\langle X_1, \ldots, X_n \rangle) \ | \ \sigma
\in \Sigma \}$
is polynomial as
$X_1=\ldots = X_n=0$ is always a solution.
\myqed

Both these results
identify pathological phenomena
and are not practical limitations.
In the first proof, we
introduced complexity %into symmetry breaking
by making the ordering NP-hard to decide. In the
second proof, we
introduced complexity %into symmetry breaking
by making the symmetry group NP-hard to decide.

In the next two sections, we will show that symmetry
breaking with other orderings is intractable.
Our argument uses a symmetry group whose
elements are easy to generate and which is isomorphic to the
symmetry group that interchanges rows and columns in
a matrix model. In addition, we suppose that the ordering
is polynomial to decide so cannot itself be a source
of computational complexity.

\section{Symmetry under Reformulation}
}

We will show that, under some modest assumptions,
we cannot make breaking symmetry computationally easier by
using a new ordering like the Gray code ordering.
Our argument breaks into two parts. First,
we observe how the symmetry of a problem changes
when we reformulate onto
an equivalent problem.
Second, we argue that we can map
onto an equivalent problem on which symmetry
breaking is easier.
%Therefore, breaking symmetry with a different
%ordering cannot have a lesser
%computational complexity.
%
%We begin by proving that
%reformulation maps the symmetry
%group of a problem onto an isomorphic
%symmetry group.

\begin{mytheorem}
If a set of constraints $S$ has a symmetry group
$\Sigma$, $S$ and
$T$ are equivalent sets of constraints,
$\pi$ is any bijection between solutions
of $S$ and $T$, and $\Pi \subseteq \Sigma$ then:
\begin{enumerate}
\item[(a)] $\pi \Sigma \pi^{-1}$ is
a symmetry group of $T$;
\item[(b)] $\Sigma$ and $\pi \Sigma \pi^{-1}$ are
isomorphic symmetry groups;
\item[(c)] if $\Pi$ generates $\Sigma$ then
$\pi \Pi \pi^{-1}$ generates $\pi \Sigma \pi^{-1}$.
\end{enumerate}
\end{mytheorem}
\myOmit{
\myproof
(a) Consider a solution $A$ of $T$ and %any
$\sigma \in \Sigma$. Then
$\pi^{-1}(A)$ is a solution of $S$. As $\sigma$ is a symmetry
of $S$, $\sigma(\pi^{-1}(A))$ is a solution of $S$. Hence,
$\pi(\sigma(\pi^{-1}(A)))$ is a solution of $T$. Thus,
$\pi \sigma \pi^{-1}$ is a symmetry of $T$.
Hence $\pi \Sigma \pi^{-1}$ is a symmetry group of $T$.

(b) $S$ and $T$ are equivalent sets of
constraints. Consider a solution $A$ of $S$.
Then the bijection $\pi$ maps the symmetry class
$[A]_{\Sigma}$ onto the isomorphic
symmetry class $[\pi(A)]_{\pi \Sigma \pi^{-1}}$.
Consider two symmetric solutions $B$ and $C$ from
$[A]_{\Sigma}$ where $B \neq C$.
As they are in the same symmetry class,
there exists $\sigma \in \Sigma$
with $\sigma(B)=C$.
The bijection $\pi$ maps
$B$ and $C$ onto
$\pi(B)$ and $\pi(C)$ respectively.
Consider the symmetry $\pi \sigma \pi^{-1}$ in
$\pi \Sigma \pi^{-1}$.
Now $\pi (\sigma (\pi^{-1} (\pi(B)))) =
\pi(\sigma(B)) = \pi(C)$.
Thus $\pi(B)$ and $\pi(C)$ are
in the same symmetry class.
As $\pi$ is a bijection and $B \neq C$,
it follows that $\pi(B) \neq \pi(C)$.
Hence, this symmetry class has the same size as
$[A]_{\Sigma}$. Thus $\Sigma$ and $\pi \Sigma \pi^{-1}$ are
isomorphic symmetry groups.

(c) Consider %any
$\sigma \in \Sigma$. There exist
$\sigma_1$, \ldots, $\sigma_n \in \Pi$ such that
$\sigma$ is generated from the product $\sigma_1 \cdots \sigma_n$.
Consider $\pi \sigma_1 \pi^{-1}$, \ldots, $\pi \sigma_n \pi^{-1} \in
\pi \Pi \pi^{-1}$. Their product is
$\pi \sigma_1 \pi^{-1}  \pi \sigma_2 \pi^{-1} \cdots
\pi \sigma_n \pi^{-1}$ which simplifies to
$\pi \sigma_1 \cdots \sigma_n \pi^{-1}$ and
thus to $\pi \sigma \pi^{-1}$. Hence $\Pi$
generates $\pi \Sigma \pi^{-1}$.
\myqed
}

\myOmit{
We next show that a sound (complete) set of
symmetry breaking constraints will be mapped
by the reformulation onto a sound (complete) set of
symmetry breaking constraints for the reformulated
problem.
\begin{mytheorem}
If a set of constraints $S$ has a symmetry group
$\Sigma$,
$B$ is a sound (complete) set of symmetry
breaking constraints for $S$,
and $S$ and $T$ are equivalent sets of constraints,
then $\pi(B)$ is a sound (complete) set of symmetry
breaking constraints for $T$
where $\pi$ is a bijection between solutions
of $S$ and $T$.
\end{mytheorem}
\myproof
%(Soundness)
Suppose $B$ is a sound set of symmetry
breaking constraints for $S$.
Consider $A \in sol(S \cup B)$.
Now $A \in sol(S)$
and $A \in sol(B)$.
But as $\pi$ is a bijection between assignments
of $S$ and $T$
$\pi(A) \in sol(T)$.
Since $A \in sol(B)$,
it follows that $\pi(A) \in sol(\pi(B))$ \cite{kwecai10}.
Thus, $\pi(A) \in sol(T \cup \pi(B))$.
Hence, there is at least one solution left by
$\pi(B)$ in every symmetry class of $T$.
That is, $\pi(B)$
is a sound set of symmetry breaking constraints.
Completeness is proved in a similar way.
%
%(Completeness)
%Suppose $B$ is a complete set of symmetry
%breaking constraints for $S$.
%Consider any $A \in sol(T \cup \pi(B))$
%Now $A \in sol(T)$ and $A \in sol(\pi(B))$.
%But as $\pi$ is a bijection between solutions of $S$
%and $T$, $\pi^{-1}$ is a bijection between solutions of
%$T$ and $S$. Hence $\pi^{â1}(A) \in sol(S)$.
%Since $A \in sol(\pi(B))$, it follows
%that $\pi^{â1}(A) \in sol(B)$.
%Thus $\pi^{â1}(A) \in sol(S \cup B)$.
%Hence, there is at most one solution left by
%$\pi(B)$ in every symmetry class of $S$.
%That is, $\pi(B)$
%is a complete set of symmetry breaking constraints.
\myqed

We have shown that reformulating onto an equivalent problem just
maps the symmetries onto isomorphic symmetries and
simply requires the same reformulation
of any symmetry breaking constraints. }
We will use this proposition to argue
that symmetry breaking with any ordering
besides the lexicographical ordering is
intractable. % for a symmetry group isomorphic to
%the group that permutes rows and columns
%in a matrix model.
%
%
%\section{Breaking Symmetry in General}
%
%Suppose we break symmetry using some other
%ordering than the usual lexicographical ordering
%on assignments. For example, suppose we
%break symmetry by insisting that any solution
%is the smallest symmetric solution in each
%symmetry class under the Gray code ordering.
%Under modest assumptions, we can lower bound the
%complexity of symmetry breaking.
%
We consider only %those
%orderings which satisfy some basic properties.
%We call these
{\em simple} orderings. In a simple ordering,
we can compute the position of any assignment in
the ordering in polynomial time, and given
any position in the ordering we can compute the
assignment at this position \ninacpp{in polynomial time}.
\myOmit{
For example, %for 0/1 variables and a lexicographical
%ordering, $\langle 0,\ldots,0,0,0 \rangle$ is
%in first position in the lexicographical ordering,
%$\langle 0,\ldots,0,0,1 \rangle$ is in
%second position,
%$\langle 0,\ldots,0,1,0 \rangle$ is in
%third position,
%and $\langle 1,\ldots,1,1 \rangle$ is in
%$2^n$th (or last) position.
%Given any assignment,
%we can compute its position in the lexicographical
%ordering in polynomial time.
%Similarly, we can compute the $k$th assignment
%in the lexicographical ordering in polynomial time.
%As a second example
%for %0/1 variables and
%the Gray code
%ordering,
$\langle 0,\ldots,0,0,0 \rangle$ is
in first position in the Gray code ordering,
$\langle 0,\ldots,0,0,1 \rangle$ is in
second position,
$\langle 0,\ldots,0,1,1 \rangle$ is
in third position,
$\langle 1,0,\ldots,0,0 \rangle$ is
in
%$2^n$th (or
last position, and
we can compute these positions in polynomial time.
Similarly, we can compute the 0$k$th assignment
in the Gray code ordering in polynomial time.
%As a third example, the Snake-Lex ordering is simple.
%
}
We now
give our main result. % which
%generalizes the result in \cite{clgrkr96}.
%that computing the Lex-Leader assignment is
%NP-hard. We prove that
%Computing the smallest
%symmetry of an assignment for {\em any} simple ordering is NP-hard.

\begin{mytheorem}
Given any simple ordering $\preceq$,
there exists a symmetry group
such that deciding if
an assignment is smallest in its symmetry
class according to $\preceq$ is NP-hard.
\end{mytheorem}
\myproof
Deciding if
an assignment is smallest in its symmetry
class according to $\leq_{\rm lex}$ is NP-hard
\cite{clgrkr96}.
Since $\preceq$ and $\leq_{\rm lex}$ are both
simple orderings, there exist polynomial
functions $f$ to map assignments onto
positions in the $\leq_{\rm lex}$ ordering,
and $g$ to map
positions in the $\preceq$ ordering to assignments.
Consider the mapping $\pi$
defined by $\pi(A)=g(f(A))$. % for any complete assignment $A$.
Now $\pi$ is a permutation that
is polynomial to compute which
maps the total ordering of assignments of
$\leq_{\rm lex}$ onto
that for $\preceq$.
Similarly,
$\pi^{-1}$ is a permutation that
is polynomial to compute which
maps the total ordering of assignments of
$\preceq$ onto
that for $\leq_{\rm lex}$.
Let $\Sigma_{rc}$ be the row and column symmetry group.
By Theorem 2,
the problem of finding
the lexicographical least
element of each symmetry class
for $\Sigma_{rc}$ is equivalent
to the problem of finding
the least
element of each symmetry class
according to $\preceq$
for $\pi \Sigma_{rc} \pi^{-1}$.
Thus,
for the symmetry group
$\pi \Sigma_{rc} \pi^{-1}$
deciding if
an assignment is smallest in its symmetry
class according to $\preceq$ is NP-hard.
\myqed

It follows that %even
\ninacpp{there exists an infinite family of symmetry groups such that}
checking a constraint
which is only satisfied by the smallest
member of each symmetry class %according to $\preceq$
is NP-hard. Note that the Gray code and Snake-Lex orderings
are simple. Hence, %a corollary of
%this proposition is that
breaking symmetry with
either ordering is NP-hard \ninacpp{for some symmetry groups}.
Note that we are not claiming
that deciding if an assignment is
smallest in its symmetry class is NP-complete.
First, we would need to worry about the size of the
input (since we are considering the
much larger class of symmetries that
act on complete assignments rather than
on literals). Second, to
decide that an assignment is the
smallest, we are also answering a complement
problem (there is {\em no} smaller symmetric
assignment). This will take us to DP-completeness
or above.

\section{Breaking Matrix Symmetry}

We next consider a common type of symmetry.
\nina{In many models, we have a matrix of decision
variables in which the rows and columns
are interchangeable \cite{matrix2001,matrix2002,ffhkmpwcp2002}.
We will show that breaking row and column symmetry specifically
is intractable with the Gray code and the Snake-Lex
orderings, as it is with the
lexicographical ordering that considers the variables in
a row-wise order.}

%We next show that a common type of symmetry
%is intractable to break when we use the Gray code or Snake-Lex orderings.
%In many models, we have a matrix of decision
%variables in which the rows and columns
%are interchangeable \cite{ffhkmpwcp2002}.
%Breaking such symmetry is intractable
%with the Gray code and the Snake-Lex
%orderings, as it is with the lexicographical ordering.
% Such matrix symmetry is also
%intractable to break with the simple
%lexicographical ordering.

\begin{mytheorem}
		\label{l:l1}
Finding the smallest solution up to row and column
symmetry for the Snake-Lex
ordering is NP-hard.
		\end{mytheorem}		
		\myproof
        We reduce from the problem of finding the Lex-Leader solution of a matrix $B$.
Let $B$ be an $n \times m $ matrix of Boolean values.
W.l.o.g. we assume $B$ does not contain a row of only ones
since any such row can be placed at the bottom of the matrix.
We embed $B$ in the matrix $M$ such that
finding $\sigma(M)$, denoted $M'$, the smallest
row and column symmetry of $M$ in the Snake-Lex ordering is equivalent
to finding the Lex-Leader of $B$.
{We ensure
        that even rows in  the Snake-Lex smallest symmetric solution of $M$ are
        taken by dummy identical rows. Then in odd rows,
%in the
%       Snake-Lex smallest symmetry of $M$,
where Snake-Lex moves from the
        left to the right along a row like Lex does, we embed the Lex-Leader solution of $B$.}

\begin{figure*}[htb]
  \centering
%\subfigure[]
{      \includegraphics[width=1\textwidth]{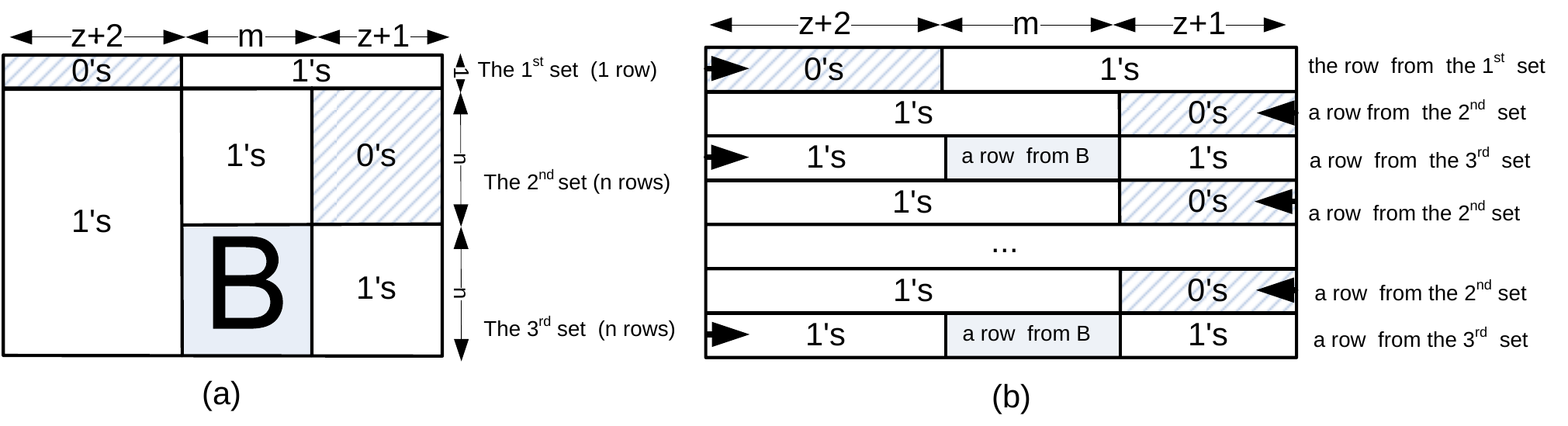} }
  \caption{(a) Construction of $M$ (b) Partial construction of $M'$. The first and all even rows are fixed.}
  \label{f:construction0}
\end{figure*}

%\begin{figure}[htb]
%  \centering
%\subfigure[]
%{      \includegraphics[width=0.5\textwidth]{Visio-snake_00.pdf} }
%  \caption{Construction of $M'$}
%  \label{f:construction00}
%\end{figure}

Let $z$ be the maximum number of zeros in any row of $B$.
We construct $M$ with $2n+1$ rows and $(z+2)+(z+1)+m$ columns so that
\nina{it contains three sets of rows.
The first set consists of a single row that contains $z+2$ zeros followed by $(z+1)+m$ ones.
The second set contains $n$ identical rows with $z+2+m$ ones followed by
$z+1$ zeros in each row.
The third set of rows contains $n$ rows such that at the $i$th row the first $(z+2)$ positions are ones,
the next $m$ positions are the $i$th row from $B$ and the last $z+1$ positions are ones again.
%it embeds the original matrix $B$
%and, additionally, contains two sets of dummy rows.
Schematic representation of $M$ is shown at Figure~\ref{f:construction0}(a).
%We let $z_1 = z+1$ and $z_2 = z+2$.
}

%
%To describe the structure of $M$, we use the notation $M[a:b;c:d]$ to denote the
%        submatrix containing values from the intersection of
%        rows from $a$ to $b$ and columns from $c$ to $d$. We let $z_1 = z+1$ and $z_2 = z+2$
%        \begin{description}
%          \item[Block 1:] $M[1:1;1:z_2]$ contains $z+2$ zeros and $M[1:1;z_2+1:z_2+z_1+m]$ contains ones.
%          \item[Block 2:] $M[2:2n+1;1:z_2]$ are set to ones.
%          \item[Block 3:] $M[2:n+1;z_2+1:z_2+z_1]$ are set to zeros and $M[2:n+1; z_2+z_1+1:z_2+z_1+m]$ are set to ones.
%          \item[Block 4:] $M[n+2:2n+1;z_2+1:z_2+z_1]$ are set to ones.
%          \item[Block 5:] $M[n+2:2n+1;z_2+1+z_1+1:z_2+z_1+m] =  B$.
%        \end{description}
%
%        Figure~\ref{f:construction0} shows the construction of $M$.

%		\begin{figure}[htb]
%  \centering
%%\subfigure[]
%{      \includegraphics[width=0.5\textwidth]{Visio-snake_1.pdf} }
%  \caption{ Partial construction of $M'$. The first and all even rows are fixed.}
%  \label{f:construction1}
%\end{figure}
\nina{ %  $M'$ that is the smallest solution up to row and column
% symmetry for the Snake-Lex ordering.
We determine positions of rows and columns that must be fixed in $M'$
up to permutation of identical rows and columns.}
        The first row of $M$ has to be the first row of
        $M'$ as no other row contains $z+2$ zeros.
        Note that this also fixes the position of columns from 1 to $z+2$ in $M$ to be the
        first columns in the  $M'$. Note also that these columns are identical
        and each of them contains the zero in the first row only.

        \nina{One of the rows in the second set  has to be the second row of  $M'$,  as none of the rows
        that embed  rows from $B$ contains $z+1$ zeros.}
        As we move from the right to the left on even rows,
        this also makes sure that last $z+1$ columns from $M$ must
        be the last columns in $M'$.
        \nina{We summarise that at this point the first two rows
        are fixed  and
        the first $z+2$ columns and the last $z+1$ columns
        in  $M'$  must be equal to a permutation of
        the first $z+2$ identical columns and the last $z+1$ identical columns  in $M$, respectively.
       }
        %Hence, these columns are moved
%        to positions from $z_2+z_1+1$
%        to  $z_2+z_1+n$.
%  Figure~\ref{f:construction00} shows the construction
%        after the first two rows are fixed in $M'$.

        By assumption,  $B$ does not contain rows with all ones.
        \nina{ Moreover, only rows that embed rows from $B$ can have the value zero
        at columns from $(z+2)+1$ to $(z+2)+m$ in $M'$.
        Hence, a row from the third set that embeds a row from $B$ has to be the third row in $M'$.
        We do not specify which row it is at this point.
        The fourth row has to be again a row from the second set as any
        of remaining rows from the second set has $z+1$ zeros
        in the last $z+1$ columns in  $M'$ while any row that embeds $B$ has at most $z$ zeros.
        We can repeat this argument for the remaining rows.}
        A schematic representation
        of the positions of rows from the first and second sets are shown in Figure~\ref{f:construction0}(b).
        \nina{Note that the first and all even rows in $M'$ are fixed.
        The only part of $M'$ not yet specified is the ordering of odd rows of $m$ columns from $(z+2)+1$ to $(z+2)+m$. These are exactly all rows from $B$.
        Hence, finding $M'$ is reduced to ordering of this set of rows and columns that embed $B$.
        Now, all columns from $(z+2)+1$ to $(z+2)+m$ are interchangeable, all odd rows except the first are interchangeable, and all elements of $M'$ except elements of $B$ are fixed by construction.
As the Snake-Lex ordering goes
        from the left to the right on odd rows like the Lex ordering,
        finding $M'$ is equivalent to finding the Lex-Leader of $B$ }
		\myqed

To show that finding the smallest row and column
symmetry in the
Gray code ordering is NP-hard, we need a technical lemma
about cloning columns in a matrix. \ninacpp{We use rowwise ordering in a matrix.}
%Consider a $n \times m $  matrix $B$ of Boolean values.
Suppose we clone each column in a $n \times m$ Boolean matrix
$B$ to give the matrix $B^c$.
Let $B^c_{gl}$ be the smallest row and
column symmetry of $B^c$ in the Gray code ordering.
%W.l.o.g. we assume that the original column precedes its clone in $B^c_{gl}$
%to make the proof simpler.

\begin{mylemma}
		\label{l:l2}
        Any original column of $B$ is followed
        by its clone %column
in $B^c_{gl}$ ignoring
permutation of identical original columns.
		\end{mylemma}		
\myproof
By  contradiction. Suppose there exists an element $B^c_{gl}[j,i+1]$
such that the original column $i$ and the next column $i+1$ are different
at the $j$th row.  We denote by  $k$ the $[j,i+1]$ element of $B^c_{gl}$
in its row-wise linearization. We ignore the rows from $j+1$ to $n$ at this point
as they are not relevant to this discrepancy.

Each pair of columns coincide on the first $j$ rows
for the first $i-1$ columns and
on the first $j-1$ rows
for the columns from $i$ to $m$.
We conclude that (1) $i$ is odd and $i+1$ is even;
(2) the number of ones between the first and the $(k-2)$th positions in the linearization of $B^c_{gl}$
  is even as each value is duplicated;
(3) the clone of the $i$th column cannot be among the first $i-1$ columns as
each such column is followed by its clone
%in the first $i-1$ columns
by assumption. %}
Hence, the clone of the $i$th column is among columns from $i+2$ to $m$.

Suppose the clone of the $i$th column is the $p$th column. Note that
the $p$th column must coincide with the $i+1$th
column at the first $j-1$ rows.
We consider two cases.
In the first case, $B^c_{gl}[j,i] = 1$ and $B^c_{gl}[j,i+1]=0$.
Note that the total number of ones at
the positions from $1$ to $k-1$  is odd
as we have one in the position $k-1$ and \nina{the number of ones in the first $k-2$
positions is even}.
Next we swap the $(i+1)$th and  $p$th columns in $B^c_{gl}$.
This will not change the first $k-1$ elements
in the linearization \nina{as the $p$th column must coincide with the $i+1$th
column at the first $j-1$ rows.}
Moreover, this swap puts %the value
$1$ in position
$k$. As the number of ones
up to the $(k-1)$th position is odd then %the value
$1$ goes
before %the value
$0$ at position $k$ in the Gray code ordering.
Hence, by swapping the $(i+1)$th and $p$th columns
we obtain a matrix that is
smaller than $B^c_{gl}$ \nina{in the Gray code ordering.} This
is a contradiction.
In the second case, $B^c_{gl}[j,i] = 0$ and $B^c_{gl}[j,i+1]=1$.
Note that the total number of ones at positions
%from
1 to $k-1$ in the linearization is even
as we have zero at the position $k-1$ and \nina{the number of ones in the first $k-2$
positions is even}.
Therefore, %the value
$0$ precedes %the value
$1$ at
position $k$ in the Gray code ordering.
By swapping  the $(i+1)$th and $p$th columns we obtain
a matrix that is smaller than $B^c_{gl}$ \nina{in the Gray code ordering}
as %the value
$0$ appears at the position $k$ instead of %the value
$1$.
This is a contradiction.
		\myqed

\begin{mytheorem}
		\label{l:l3}
Finding the smallest solution up to row and column
symmetry for the Gray code
ordering is NP-hard.
 		\end{mytheorem}		
\myproof
We again reduce from the problem of finding the Lex-Leader solution of a matrix $B$.
We clone every column of $B$ and obtain a new matrix $B^c$.
Let $B^c_{gl}$ be the smallest row and column symmetry of
$B^c$ in the Gray code ordering.
Lemma~\ref{l:l2} shows that each original column is followed
by its clone in $B^c_{gl}$. Next we delete all
clones by removing every second column. We call the resulting
matrix $B_{l}$.
We prove that $B_{l}$ is the Lex-Leader of $B$ by contradiction.
Suppose there exists a matrix $M$ which
is the Lex-Leader of $B$ %and this matrix
that is different from $B_l$.
Hence, $M$ is also the Lex-Leader of $B_l$.
We find the first element $M[j,i]$
where $B_l[j,i] \neq M[j,i]$ \nina{in the row-wise linearization of $M$ and $B_l$}, so that
$B_l[j,i] = 1$ and $M[j,i] =0$.
We denote by  $k$  the position of the $[j,i]$ element of $M$ in its row-wise linearization.
We clone each column of $M$ once and put  each cloned column
right after its original column.  We obtain a new matrix $M^c$.
We show that $M^c$ is smaller than $B^c_{gl}$ in the Gray code
ordering to obtain
a contradiction.

As $B_l[j,i] = 1$ and $M[j,i] =0$ then
$B^c_{gl}[j,2i-1] = 1$ and $M^c[j,2i-1] =0$ { because
the matrices $B^c_{gl}$ and $M^c$
are obtained from $B_l$ and $M$ by cloning each
column and putting each clone right after its original column.}
As  $B_l$ and $M$ coincide on the first  $k-1$ positions
then $B^c_{gl}$ and $M^c$ coincide in the first  $2k-2$ positions.
%Another important point is that the first difference occurs at position $2k-1$
%which is odd as we duplicated each value in positions from $1$ to $k-1$.
By transforming $B_l$ and $M$ to $B^c_{gl}$ and $M^c$,
we duplicated each value in positions from $1$ to $k-1$.
Hence, the total number of ones in positions from $1$ to $2k-2$ in $B^c_{gl}[j,i]$ and $M^c[j,i]$ is even.
Therefore, the value zero precedes the value one at
position $2k-1$ in the Gray code ordering.
By assumption, the value in the position $2k-1$ in $B^c_{gl}$, which is $B^c_{gl}[j,2i-1]$,
is 1, and the position $2k-1$ in $M^c$, which is $M^c[j,2i-1]$, is $0$. Hence, $M^c$ is smaller than $B^c_{gl}$ in the Gray code ordering.
\myqed

We conjecture that row and column symmetry will be
intractable to break for other simple orderings. However,
each such ordering may require a new proof. 

\section{Other Symmetry Breaking Constraints}

Despite these negative theoretical results,
there is still the possibility for other
orderings on assignments
to be useful when breaking symmetry in practice.
It is interesting therefore to develop propagation algorithms
for different orderings.
Propagation algorithms are used to prune the search
space by enforcing properties like domain
consistency. A constraint is \emph{domain consistent} (\emph{DC})
iff when a variable is assigned any value in its domain, there
exist compatible values in the domains of the other variables.
% of the constraint.

\subsection{Gray Code Constraint}

We give an efficient encoding for the \ninacp{new} global constraint
$Gray([X_1,\ldots,X_n], [Y_1,\ldots,Y_n])$
that ensures $\langle X_1,\ldots,X_n \rangle$ is before
or equal in position to  $\langle
Y_1,\ldots,Y_n \rangle$
in the Gray code ordering where $X_i$ and $Y_j$ are 0/1
variables. %For a variable symmetry, we need just then
%set $Y_i = X_{\sigma(i)}$ where $\sigma$ is an appropriate
%bijection on variable indices,
%whilst for a value symmetry, we need just set
%$Y_i = \theta(X_i)$ where $\theta$ is an appropriate
%bijection on values.
%
\nina{We encode the transition relation of an automaton
with 0/1/-1 state variables, $Q_1$
to $Q_{n+1}$ that reads
a sequence $\langle X_1,Y_1,\ldots,X_n,Y_n\rangle$
and ensures that the two sequences are ordered
appropriately.
We consider the following  decomposition
}
%We suppose the existence of a sequence of 0/1/-1 state variables, $Q_1$
%to $Q_{n+1}$.
%We encode this constraint
%by means of the following decomposition
where $1 \leq i \leq n$:
\myOmit{
\begin{eqnarray*}
%&Q_1 =  1,& \\
&Q_1=1,\ \ \ Q_i \neq 1  \vee  X_i \leq Y_i,\ \ \  Q_i \neq -1 \vee  X_i \geq Y_i,
\ \ \ X_i = Y_i \vee Q_{i+1}=0, & \\
%&X_i = Y_i \vee Q_{i+1}=0,& \\
&X_i \neq 0 \vee Y_i \neq 0 \vee Q_{i+1}=Q_i, %&\\
%&
\ \ \ X_i = 0 \vee Y_i = 0 \vee Q_{i+1}=-Q_i. &
\end{eqnarray*}}
\begin{eqnarray*}
%&Q_1 =  1,& \\
&Q_1=1,\ Q_i \neq 1  \vee  X_i \leq Y_i,\ Q_i \neq -1 \vee  X_i \geq Y_i, & \\
& X_i = Y_i \vee Q_{i+1}=0, \ X_i = 0 \vee Y_i = 0 \vee Q_{i+1}=-Q_i. &
\end{eqnarray*}
%
%If we want to ensure that
%$\langle X1,\ldots,Xn \rangle$ is {\em strictly} before
%$\langle Y1,\ldots,Yn \rangle$, we just post
%the additional constraint $Q_{n+1} =  0$.
%This decomposition does not hinder propagation.

We can show that this decomposition not only preserves
the semantics of the constraint but also does
not hinder propagation.
\sloppy
\begin{mytheorem}
Unit propagation on this decomposition
enforces domain consistency
on $Gray([X_1,\ldots,X_n],[Y_1,\ldots,Y_n])$
in $O(n)$ time.
\end{mytheorem}
\myproof
(Correctness)
$Q_i=0$ as soon as the
two vectors are ordered correctly.
$Q_i=1$ iff %the $i$th bits,
$X_i$ and $Y_i$
are ordered in the Gray code
ordering with 0 before 1. $Q_i=-1$
iff the $i$th bits,
$X_i$ and $Y_i$
are ordered in the Gray code
ordering with 1 before 0.
$Q_{i+1}$ stays the same polarity
as $Q_i$
iff $X_i=Y_i=0$ and flips
polarity iff $X_i=Y_i=1$.

(Completeness)
\nina{This follows from the completeness
of CNF encoding of the corresponding automaton~\cite{qwcp07}
and the fact that unit propagation on this set of constraints
%for each $i$, $1 \leq i \leq n$,
enforces DC on a table constraint that
encodes the transition relation.
}

%is Berge acyclic. Thus
%unit propagation enforces domain consistency.
%guarantees
%the existence of a support for every value.

(Complexity)
There are $O(n)$ disjuncts in the decomposition.
Hence unit propagation takes $O(n)$ time.
In fact, it is possible to show that
the {\em total}
time to enforce DC
down a branch of the search tree is $O(n)$.
\myqed

Note that this decomposition
can be used to break symmetry with the Gray code ordering
in a SAT solver.
% Note also that this decomposition
%is well defined for non-binary domains. We shall
%need this for our third set of experiments
%where domains are 0/1/2.

\subsection{Snake-Lex Constraint}
\sloppy
%In this section we consider common
%symmetry breaking constraints are useful  in  matrix models.
For row and column symmetry, we can break symmetry with
the $\DLex$ constraint that
lexicographically orders rows and columns,
or the $\snakelex$ constraint. This
is based on the smallest row and column permutation of the
matrix according to an ordering on assignments
that linearizes the matrix in a snake-like manner \cite{snakelex}.
The (columnwise) $\snakelex$ constraint can be enforced
by a conjunction of $2m-1$ lexicographical
ordering constraints on pairs
of columns and $n-1$ lexicographical
constraints on pairs of intertwined rows.
\ninacpp{To obtain the rowwise $\snakelex$ constraint,
we transpose the matrix and then order as in the columnwise $\snakelex$.}
Note that $\DLex$ and $\snakelex$ \emph{only break a subset} of
the row and colum symmetries.
However, they are very useful in practice.
It was shown in~\cite{knwcp10}, that enforcing
\emph{DC} on the $\DLex$  constraint is NP-hard. Hence
we typically decompose it into separate row and column
constraints.
Here, we show that enforcing \emph{DC} on the $\snakelex$ constraint is
also NP-hard. It is therefore also
reasonable to propagate $\snakelex$ by
decomposition.
%
%We recall definitions of these constraints.
%Consider a matrix of variable $B_{n\times m}$.
%For simplicity and WLOG, we consider a $n$ by $m$ matrix
%of variables with $n$ and $m$ even.
%The $\DLex$ constraint consists of $n-1$ row ordering
%constraints,
%$ X_{i,1}\ldots X_{i,m}    \leq   X_{i+1,1} \ldots  X_{i+1,m}$,
%$i=1,\ldots,n-1$ and $m-1$ column ordering
%constraints,
%$ X_{1,i}\ldots X_{m,i}    \leq   X_{1,i+1} \ldots  X_{m,i+1}$,
%$i=1,\ldots,m-1$. A solution of the constraint satisfies
%%all these constraints. Hence, the rows and columns of a matrix $B$ is lexicographically
%ordered.
%
\myOmit{
The (columnwise) $\snakelex$ constraint ensures a conjunction of $2m-1$ ordering constraints on columns and $n-1$ constraints on rows. }
\myOmit{We
denote constraints on columns  $c_{i,j}, i=1,\ldots,m-1, j \in \{i+1,\min(i+2,m)\}$,
and constraints on rows $r_i$, $i=1,\ldots,n-1$:

\noindent
\mbox{\scriptsize
$
\begin{array}{|lr|c|lr|}
\cline{1-2}
\cline{4-5}
  \multicolumn{2}{|c|}{\textrm{Constraints on columns} } &\ &    \multicolumn{2}{c|}{\textrm{Constraints on rows}}\\
\cline{1-2}
\cline{4-5}
   c_{1,2}: & X_{1,1} ..  X_{n,1}    \leq   X_{1,2}  ..  X_{n,2} &\ & r_1: & X_{1,1} X_{2,2} X_{1,3}..  X_{2,m}    \leq   X_{2,1}, X_{1,2} X_{2,3} ..  X_{1,m} \\
  c_{1,3}: &X_{1,1}  ..  X_{n,1}    \leq   X_{1,3} ..  X_{n,3} & \   & r_2: & X_{2,1} X_{3,2} X_{2,3}..  X_{3,m}    \leq   X_{3,1} X_{2,2} X_{3,3} ..  X_{2,m} \\
  c_{2,3}: &X_{n,2} ..  X_{1,2}    \leq  X_{n,3} ..  X_{1,3}&  \    & r_3: & X_{3,1} X_{4,2} X_{3,3}..  X_{4,m}    \leq   X_{4,1} X_{3,2} X_{4,3} ..  X_{3,m} \\
   c_{2,4}: &X_{n,2} ..  X_{1,2}    \leq  X_{n,4}  ..  X_{1,4} & \  & r_4: & X_{4,1} X_{5,2} X_{4,3}..  X_{5,m}    \leq   X_{5,1} X_{4,2} X_{5,3} ..  X_{6,m}\\
  & \ldots  & \   &  &  \ldots \\
  c_{m-1,m}: & X_{1,m-1} .. X_{n,m-1}   \leq  X_{1,m} .. X_{n,m}& \  & r_{n-1}: & X_{n-1,1} X_{n,2} ..  X_{n,m}    \leq   X_{n,1}X_{n-1,2} ..  X_{n-1,m}\\
\cline{1-2}
\cline{4-5}
\end{array}
$
}
%
%\begin{eqnarray*}
%
% c_{1,2} & X_{1,1}, X_{2,1}, \ldots,  X_{n,1}    \leq_{lex}   X_{1,2}, X_{2,2}, \ldots,  X_{n,2} \\
% c_{1,3} &X_{1,1}, X_{2,1}, \ldots,  X_{n,1}    \leq_{lex}   X_{1,3}, X_{2,3}, \ldots,  X_{n,3} \\
% c_{2,3} &X_{n,2}, X_{n-1,2}, \ldots,  X_{1,2}    \leq_{lex}  X_{n,3}, X_{n-1,3}, \ldots,  X_{1,3} \\
% c_{2,4} &X_{n,2}, X_{n-1,2}, \ldots,  X_{1,2}    \leq_{lex}  X_{n,4}, X_{n-1,4}, \ldots,  X_{1,4} \\
%    &    \ldots    \\
% c_{m-1,m} & X_{1,m-1}, X_{2,m-1}, \ldots,  X_{n,m-1}   \leq_{lex}  X_{1,m}, X_{2,m}, \ldots,  X_{n,m}
%
%\end{eqnarray*}
%The row constraints  are
%
%\begin{eqnarray*}
%r_1 & X_{1,1}, X_{2,2}, X_{1,3}\ldots,  X_{2,m}    \leq_{lex}   X_{2,1}, X_{1,2}, X_{2,3} \ldots,  X_{1,m} \\
%r_2 & X_{2,1}, X_{3,2}, X_{2,3}\ldots,  X_{3,m}    \leq_{lex}   X_{3,1}, X_{2,2}, X_{3,3} \ldots,  X_{2,m} \\
%    &   \ldots    \\
%r_{n-1} & X_{n-1,1}, \ldots,  X_{n,m}    \leq_{lex}   X_{n,1},  \ldots,  X_{n-1,m} \\
%\end{eqnarray*}
}

\begin{mytheorem}
\label{t:snake_double}
Enforcing %domain consistency
DC on the  $\snakelex$ constraint is NP-hard.
\end{mytheorem}
\myproof
%(Sketch) \ninacpp{A full proof is in the online technical report. %~\cite{}.
%Let $X$ be a $n$ by $m$ matrix of Boolean variables. %, $B_{n\times m} = \{X_{i,j}\}$.
%The main idea is to embed $X$ in to a specially constructed  matrix in such a way
%that  enforcing \emph{DC} on the  $\DLex$ constraint on $X$ (which
%we already know is NP-hard) is equivalent to
%enforcing \emph{DC} on the  $\snakelex$ constraint on this larger matrix.}
%We prove there exists
%a $n+2$ by $2m$ matrix $M$ such that
%enforcing \emph{DC} on the  $\DLex$ constraint on $X$ (which
%we already know is NP-hard) is equivalent to
%enforcing \emph{DC} on the  $\snakelex$ constraint on $M$.
%We illustrate the construction of $M$ for
%$n=m=6$ in the following figure:
%\begin{center}
%%  \centering
%      \includegraphics[width=0.4\textwidth]{snake_dlex.pdf}
%\end{center}
%\myOmit{
{
Let $B$ be a $n$ by $m$ matrix of Boolean variables, $B_{n\times m} = \{X_{i,j}\}$.
The main idea is to embed $B$ in to a specially constructed  matrix $M$ in such a way
that  enforcing \emph{DC} on the  $\DLex$ constraint on $B$ (which
we already know is NP-hard) is equivalent to
enforcing \emph{DC} on the  $\snakelex$ constraint on $M$.
We transform the matrix of variables $B$ into a new matrix
of variables $M$ with $n+2$ rows, $v=1,\ldots,n+2$ and $2m$ columns, $k=1,\ldots,2m$.

\begin{displaymath}
M[v,k] = \left\{ \begin{array}{ll}
X_{v-1,(k+1)/2} & \textrm{if $v \in [2,n+1]$ and $k$ is odd} \\
1 & \textrm{if $v \in [2,n+1]$ and $k$ is even } \\
0 & \textrm{if $v=1$ and $k$ is odd or if $v=n+2$ and $k$ is even} \\
1 & \textrm{if $v=1$ and $k$ is even or if $v=n+2$ and $k$ is odd} \\
\end{array} \right.
\end{displaymath}

Figure~\ref{f:snake_dlex} shows the transformation for a matrix $B$
with $n=6$ and $m = 6$.

\emph{Row constraints on $M$.} The first constraint $r_1$ in $M$ is
$0, 1, 0, \ldots,  0  \leq   X_{1,1}, 1,  X_{1,2}, \ldots,  X_{1,m}$
which can be simplified to an entailed constraint
$0,\ldots,0    \leq   X_{1,1},  X_{1,2},  X_{2,3} \ldots,  X_{1,m}$.  Similarly,
the constraint $r_{n+2}$ is entailed.
\begin{wrapfigure}[12]{l}{6cm}
  \centering
      \includegraphics[width=0.4\textwidth]{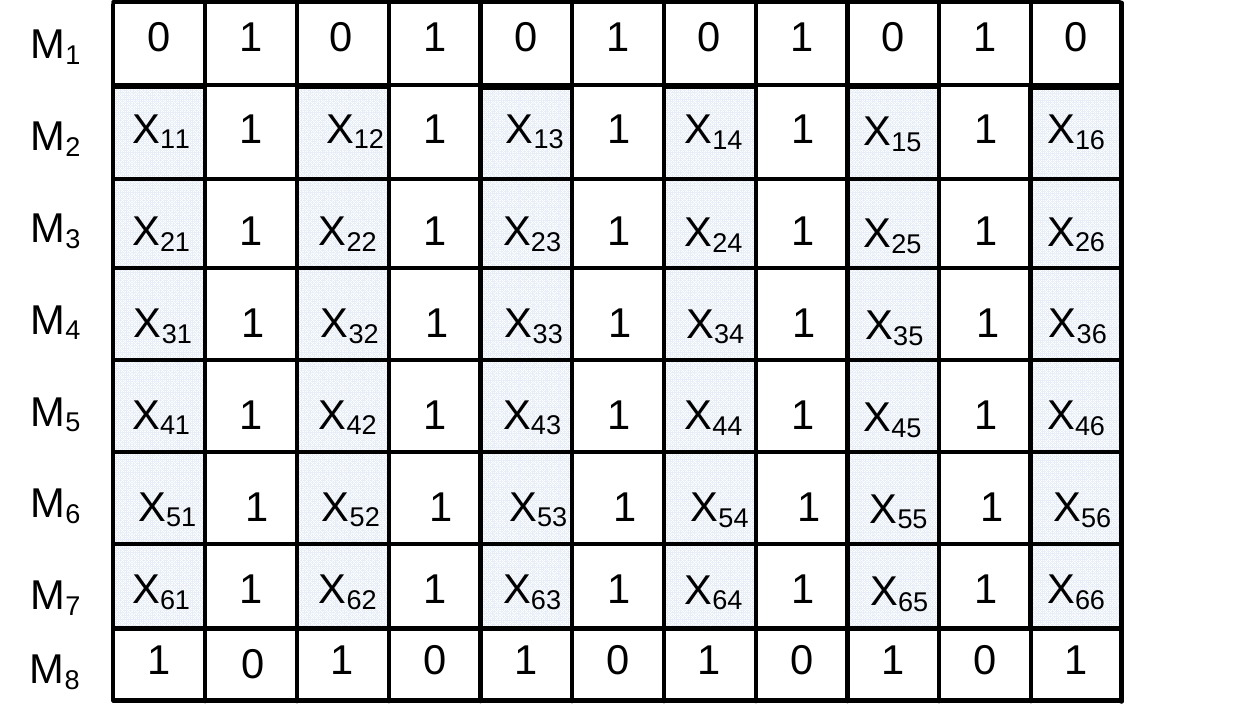}
  \caption{Construction of $M$ in the proof of Proposition 7.}
  \label{f:snake_dlex}
\end{wrapfigure}
Consider the $r_v$ constraint, $v \in [2,n]$ in $M$:
$X_{v-1,1}, 1 , X_{v-1,2}, 1, X_{v-1,3},\ldots,  X_{v-1,m}   \leq  $ $  X_{v,1}, 1, X_{v,2}, 1, X_{v,3}, \ldots,  X_{v,m}$.
As all even elements coincide we can simplify the constraint to
$X_{i,1},  X_{i,2}, X_{i,3},\ldots,  X_{i,m}    \leq   X_{i+1,1}, X_{i+1,2}, X_{i+1,3}, \ldots,  X_{i+1,m}$,
$i = v-1 $ so that $ i\in[1,2,\ldots,n-1]$.
Hence, all row constraints are simplified to the row ordering constraints of the $\DLex$ constraint.

\emph{Column constraints on $M$.}
The constraints $c_{k,k+1}$, $k \in \{1,3,\ldots,2m-1\}$, are entailed as
$0, X_{1,(k+1)/2}, X_{2,(k+1)/2}, \ldots,  X_{n,(k+1)/2}, 1     \leq    1, 1, \ldots,  0$.
The constraints $c_{k,k+1}$ and $c_{k,k+2}$, $k \in \{2,4,\ldots,2m\}$, are entailed as
$c_{k,k+1}$  is
$0, 1, \ldots,  1     \leq   1, X_{n,k/2+1}, X_{n-1,k/2+1}, \ldots,  X_{1,k/2+1}, 1$
and $c_{k,k+2}$ is $0, 1, \ldots,  1     \leq   0, 1, \ldots,  1 $.
Therefore, the only not entailed constraints are
$c_{k,k+2}$, $k \in \{1,3,\ldots,2m-1\}$:
$0, X_{1,(k+1)/2}, X_{2,(k+1)/2}, \ldots,  X_{n,(k+1)/2}, 1    \leq   1, X_{1,(k+1)/2+1}, X_{2,(k+1)/2+1}, \ldots,  X_{n,(k+1)/2+1}, 1$.
These constraints can be simplified to
$X_{1,j}, X_{2,j}, \ldots,  X_{n,j}    \leq   X_{1,j+1}, X_{2,j+1}, \ldots,  X_{n,j+1}$,
$j = (k+1)/2$ so that $j\in \{1,2,\ldots,m-1\}$ which
gives the column ordering of the $\DLex$ constraint.
Hence, enforcing \emph{DC} on the $\snakelex$ constraint  on $M$
enforces \emph{DC} on the $\DLex$ constraint  on $B$.}
\myqed

\section{Experimental Results}

%To explore the utility of breaking
%symmetry with other %total
%orderings,
%on assignments,

%In each domain, we compared a wide range of symmetry
%breaking constraints with the
%default branching heuristic, and
%then explored the impact of
%branching heuristics on those symmetry breaking
%constraints that were effective with the
%default branching heuristic.
We tested two hypotheses that provide advice to
the modeller when breaking symmetry.
\begin{enumerate}
\item other orderings besides the
lexicographical ordered can be effective when breaking symmetry
in practice;
\item symmetry breaking %constraints
should
align with the branching heuristic,
and with the objective function.
\end{enumerate}
All our experiments report the time
to find an optimal solution {\em and} prove it optimal. We believe
that optimisation is often a more realistic setting
in which to illustrate the practical benefits
of symmetry breaking, than satisfaction experiments
which either find one or all solutions.
Breaking symmetry in optimisation
problems is important as we must traverse
the whole search space when proving optimality.
All our experiments used the BProlog 7.7
constraint solver.
This solver
%was ranked top in two categories
%in the International Competition of CSP Solvers, and
took second place in the ASP 2011 solver
competition.
The three sets of experiments took around
one CPU month on a MacBook Pro with an Intel Core i5
2 core 2.53 GHz processor, with
%256 KB L2 cache/core, 3 MB L3 and
4GB of memory.
%The three experiments used three
%different benchmark problems: maximum density still life,
%low autocorrelation binary sequences, and peaceable armies
%of queens.
The three domains were chosen as
representative of optimisation
problems previously studied in symmetry
breaking. We observed similar results in these
as well as other domains.

\subsection{Maximum density still life problem}

This is prob032 in CSPLib \cite{csplib}.
This problem arises in Conway's
Game of Life, and was popularized by Martin Gardner.
Given a $n$ by $n$ submatrix of the
infinite plane, we want to find and prove optimal
the pattern of maximum density which does
not change from generation to generation.
{For example, an optimal solution for $n=3$ is:
$$
{%\scriptsize
\begin{array}{c|c|c|c|c}
& & & & \\ \hline
& & $\textbullet$ & $\textbullet$ &\\ \hline
& $\textbullet$ & & $\textbullet$ & \\ \hline
& $\textbullet$ & $\textbullet$ & & \\ \hline
& & & &
\end{array}
}
$$
This is a still life as every live square has between
2 and 3 live neighbours, and every dead square
does not have 3 live neighbours.}
We use the simple 0/1 constraint model from \cite{btaor04}.
This problem has the 8 symmetries of the square as
we can rotate or reflect any still life to
obtain a new one.
Bosch and Trick %(\citeyear{btaor04})
argued
that
{\em ``\ldots The symmetry embedded in this problem is very strong,
leading both to algorithmic insights and algorithmic difficulties\ldots''.}

Our first experiment
used the default search strategy to find and prove optimal
the still life of maximum density for a given
$n$. The default strategy instantiates variables row-wise
across the matrix. Our goal here is
to compare the different symmetry
breaking methods with an ``out of the box'' solver.
We then compare the impact of the branching
heuristic on symmetry breaking.
We broke symmetry with either the lexicographical
or Gray code orderings, finding
the smallest (lex, gray) or largest
(anti-lex, anti-gray) solution in each
symmetry class. In addition, we linearized
the matrix either row-wise (row), column-wise (col),
snake-wise along rows (snake), snake-wise along
columns (col-snake), or in a clockwise spiral (spiral).
Table 1 gives results for the
20 different symmetry breaking methods constructed
by using 1 of the 4 possible solution orderings and
the 5 different linearizations,
as well with no symmetry breaking (none).

\begin{table}[htbp]
{\scriptsize
\begin{center}
\begin{tabular}{|c||r|r|r|r|r|} \hline
{Symmetry breaking} & %3 &
$n=4$ & 5 & 6 & 7 & 8 \\ \hline \hline
 none & %23 &
 176 & \ \ 1,166 & \ \ 12,205 & \ \ {231,408} & \ \ 5,867,694 \\
 gray row & %14 &
 91 & 446 & 5,702 & 123,238 & 2,507,747 \\
 anti-lex row & %12 &
 84 & 424 & 5,473 & 120,112 & 2,416,266 \\
 anti-gray col-snake & %21 &
 68 & 500 & 5,770 & 72,691 & 2,332,085 \\
 gray spiral & %16 &
 86 & 541 & 6,290 & 120,051 & 2,311,854 \\
 gray snake & %16 &
 80 & 477 & 5,595 & 120,601 & 2,264,184 \\
 anti-lex col-snake & %13 &
 79 & 660 & 4,735 & 66,371 & 2,254,325 \\
 anti-lex spiral & %16 &
 81 & 507 & 6,174 & 119,262 & 2,241,660 \\
 anti-lex col & %14
 74 & 718 & 3,980 & 68,330 & 2,215,936 \\
 anti-lex snake & %15 &
 68 & 457 & 5,379 & 117,479 & 2,206,189 \\
 lex spiral & %14 &
 48 & 434 & 4,025 & 90,289 & 2,028,624 \\
 lex col-snake & %16 &
 77 & 359 & 5,502 & 76,400 & 2,003,505 \\
 lex col &
 80 & 560 & 4,499 & 83,995 & 2,017,935 \\
 lex row & %14 &
 33 & 406 & 2,853 & 87,781 & 1,982,698 \\
 lex snake & %14 &
 35 & 407 & 2,965 & 86,331 & 1,980,498 \\
 anti-gray col & %21 &
 70 & 522 & 5,666 & 75,930 & 1,925,613 \\
 gray col & %13
 65 & 739 & 3,907 & 87,350 & 1,899,887 \\
 gray col-snake & %13 &
 62 & 693 & 3,833 & 82,736 & 1,880,506 \\
 anti-gray row & %10 &
 {\em 26} & 269 & 2,288 & 38,476 & 1,073,659 \\
 anti-gray spiral & %11 &
 27 & 279 & 2,404 & 40,224 & 1,081,006 \\
 anti-gray snake & %11 &
 28 & {\em 262} & {\em 2,203} & {\em 38,383} & {\em 1,059,704} \\
\hline
\end{tabular}
\end{center}
}
\caption{Backtracks required to find and prove
optimal the maximum density still life of size $n$ by $n$ using
the default branching heuristic. Column winner is in {\em emphasis}.
}
%\vspace{-3em}
\vspace{-1em}
\end{table}

We make some observations about these results.
First, the Lex-Leader method (lex row) is
beaten by many methods.
For example, the top three methods all use the anti-Gray code ordering.
Second, lex tends to work better than anti-lex,
but anti-gray better than gray.
We conjecture this is
because anti-gray tends to align better with the maximization
objective than gray, but anti-lex is too aggressive
as \ninacpp{ the maximum density still life can have
more dead cells than alive cells.}
Third, although we eliminate all 7 non-identity symmetries,
the best method is only about a factor 6
faster than not breaking symmetry at all.

To explore the interaction between symmetry breaking
and the branching heuristic, we report results in Table 2
using branching heuristics besides the default row-wise
variable ordering. We used the best
symmetry breaking method
for the default row-wise branching heuristic
(anti-gray snake),
the worst symmetry breaking method for
the default branching heuristic (gray row),
a standard method (lex row),
as well as no symmetry breaking (none).
We compared the default branching heuristic
(row heuristic) with branching heuristics
that instantiate variables column-wise
(col heuristic), snake-wise along rows (snake heuristic),
snake-wise along columns (col-snake heuristic),
in a clockwise spiral from top left towards
the middle (spiral-in heuristic),
in an anti-clockwise spiral from the middle
out to the top left (spiral-out heuristic),
by order of degree (degree heuristic),
and by order of the number of attached
constraints (constr heuristic).
Note that there is no value in reporting results
for domain ordering heuristics like fail-first as domains
sizes are all binary.
%all 0/1, and domain ordering heuristics degenerate
%to the appropriate tie-breaking rule.

\myOmit{
\begin{table}[htbp]
{\scriptsize
\begin{center}
\begin{tabular}{|c||r|r|r|r|} \hline
{Branching/SymBreak} & none & gray row & lex row & anti-gray snake \\ \hline \hline
constr heuristic & {\em 187,630} & 1,194,236 & 173,340 & 2,934,032 \\
spiral-out heuristic & 1,563,188 & 1,542,777 & 1,549,020 & 1,939,335 \\
spiral-in heuristic & 1,965,625 & 721,873 & 520,165 & 350,206 \\
degree heuristic & 270,935 & 78,014 & 228,280 & 254,661 \\
col heuristic & 231,408 & {\em 74,829} & {\em 63,268} & 78,055 \\
col-snake heuristic & 230,581 & 84,043 & 82,000 & 61,687 \\
snake heuristic  & 230,581 & 85,381 & 66,715 & 39,590 \\
%ff heuristic & 231,408 & 123,228 & 87,781 & {\bf 38,383} \\
row heuristic & 231,408 & 123,228 & 87,781 & {\bf 38,383} \\
\hline
\end{tabular}
\end{center}
}
\caption{Backtracks required to find the
$7$ by $7$ still life of maximum density
and prove optimality for different branching
heuristics and symmetry breaking constraints.
Overall winner is in {\bf bold} font.
%whilst the column winners are in {\em emphasis}.
%Column max and min in { bold} font.
%``-'' indicates a time-out of 10 minutes.
}
%\vspace{-3em}
\end{table}
}

\begin{table}[htbp]
{\scriptsize
\begin{center}
\begin{tabular}{|c||r|r|r|r|} \hline
{Branching/SymBreak} & none & gray row & lex row & anti-gray snake \\ \hline \hline
spiral-out heuristic & \ \ 196,906,862 & \ \ 24,762,297 &  \ \ 194,019,848 & 222,659,696 \\
spiral-in heuristic & 65,034,993 & 18,787,751 & 12,662,207 & 9,292,164 \\
constr heuristic & {\em 5,080,541} & 2,816,355 & 3,952,445 & 8,590,077\\
degree heuristic & 6,568,195 & 2,024,955 & 6,528,018 & 7,053,908 \\
col-snake heuristic & 5,903,851 & {\em 1,895,920} & 1,849,702 & 2,127,122 \\
col heuristic &  5,867,694 & 2,212,104 & {\em 1,634,016} & 1,987,864  \\
snake heuristic  & 5,903,851 & 1,868,303 & 2,043,473 & 1,371,200 \\
row heuristic & 5,867,694 & 2,507,747 & 1,982,698 & {\bf 1,059,704} \\
\hline
\end{tabular}
\end{center}
}
\caption{Backtracks required to find the
$8$ by $8$ still life of maximum density
and prove optimality for different branching
heuristics and symmetry breaking constraints.
Overall winner is in {\bf bold}. % font.
%whilst the column winners are in {\em emphasis}.
%Column max and min in { bold} font.
%``-'' indicates a time-out of 10 minutes.
}
%\vspace{-3em}
\vspace{-0.5em}
\end{table}

We make some observations about these results.
First, the symmetry breaking method with
the best overall performance (anti-gray snake + row heuristic)
had the worst performance with
a different branching heuristic (anti-gray snake + spiral-out
heuristic).
Second, we observed good performance
when the branching heuristic aligned with
the symmetry breaking (e.g. anti-gray snake + snake heuristic).
Third, a bad combination of branching heuristic
and symmetry breaking constraints (e.g.
anti-gray snake + spiral-out heuristic) was
worse than all of the branching heuristics
with no symmetry breaking constraints.
Fourth, the default row heuristic was
competitive. It %offered the best overall
%performance, and
was best or not far from best
in every column.

{
\subsection{Low autocorrelation binary sequences}

This is prob005 in CSPLib \cite{csplib}.
The goal is to find the binary sequence
of length $n$ with the lowest autocorrelation.
We used a standard model from
one of the first studies into symmetry breaking
\cite{sbds}.
This model contains a triangular
matrix of 0/1 decision variables,
in which the sum of the $k$th row equals
the $k$th autocorrelation.
Table 3 reports results to find
the sequence of lowest autocorrelation
and prove it optimal.
We used the default variable
ordering heuristic (left2right) that instantiates variables
left to right from the beginning
of the sequence to the end.
The model has 7 non-identity symmetries
which leave the autocorrelation unchanged.
We can reverse the sequence, we can
invert the bits, we can invert just the even bits,
or we can do some combination of these operations.
We broke all 7 symmetries
by posting the constraints that, within its
symmetry class, the sequence
is smallest in the lexicographical or Gray code orderings (lex, gray) or
largest (anti-lex, anti-gray).
In addition, we also considered
symmetry breaking constraints that took the variables in reverse order
from right to left (rev), alternated the variables
from both ends inwards to the middle (outside-in),
and from the middle out to both ends (inside-out).

\begin{table}[htbp]
{\scriptsize
\begin{center}
\begin{tabular}{|c||r|r|r|r|r|r|r|} \hline
{Symmetry breaking} & %3 &
$n=12$ & 14 & 16 & 18 & 20 & 22 & 24 \\ \hline \hline
none & 2,434 & \ \ 9,487 & \ \ 36,248 & \ \ 126,057 & \ \ 474,915 & \ \ 1,725,076 & \ \ 7,447,186 \\
anti-gray outside-in & 2,209 & 6,177 & 18,881 & 92,239 & 310,473 & 1,223,155 & 4,966,068 \\
gray outside-in & 1,351 & 5,040 & 19,152 & 68,272 & 350,790 & 903,441 & 4,526,114 \\
lex outside-in & 869 & 3,057 & 11,838 & 43,669 & 262,935 & 557,790 & 3,330,931 \\
gray & 704 & 2,400 & 10,158 & 36,854 & 158,080 & 468,317 & 3,048,723 \\
lex &  707 & 2,408 & 10,178 & 36,885 & 158,132 & 468,390 & 3,047,241 \\
gray rev & 699 & 1,790 & 9,892 & 25,551 & 147,911 & 329,897 & 2,706,466 \\
anti-lex outside-in & 1,262 & 2,704 & 14,059 & 67,848 & 179,219 & 544,116 & 2,579,981 \\
anti-gray  & 1,036 & 2,226 & 9,889 & 45,375 & 167,916 & 606,977 & 2,436,236 \\
anti-lex  & 1,522 & 3,087 & 10,380 & 51,162 & 281,789 & 920,543 & 2,415,736 \\
lex rev  & 634 & 1,751 & 7,601 & 23,218 & 127,438 & 299,877 & 2,160,463 \\
anti-lex rev & 549 & 1,707 & 9,398 & 32,638 & 117,367 & 398,822 & 2,092,787 \\
gray inside-out & 662 & 1,582 & 6,557 & 25,237 & 89,365 & 248,135 & 1,667,262 \\
lex inside-out & 640 & 1,549 & 6,478 & 25,049 & 88,978 & 247,558 & 1,665,054 \\
anti-gray rev & 1,007 & 1,661 & 6,894 & 29,689 & 86,198 & 312,038 & 1,422,693 \\
anti-gray inside-out & {\em 412} & 1,412 & 5,934 & 22,942 & {\em 82,673} & {\em 245,259} & 1,271,986 \\
anti-lex inside-out & 629 & {\em 1,320} & {\em 4,558} & {\em 19,811} & 138,337 & 291,050 & {\em 927,321} \\

\hline
\end{tabular}
\end{center}
}
\caption{Backtracks required to find the $n$ bit binary
sequence of lowest autocorrelation and prove optimality with the default branching heuristic.
%Column winners are in {\em emphasis}.
}
\vspace{-3em}
\end{table}

We make some observations about these results.
First, the best two symmetry breaking methods
both look at variables starting from the middle
and moving outwards to both ends (inside-out).
By comparison, symmetry breaking constraints that
reverse this ordering of variables  (outside-in)
perform poorly. We conjecture this is because the middle
bits in the sequence are more constrained, appearing
in more autocorrelations, and so are more important
to decide early in search.
Second,
% when comparing two methods, there may not be a clear
%winner. For example,
%anti-lex inside-out beats anti-gray inside-out on
%$n=14$, 16, 18 and 24, but is
%beaten on $n=12$,
%20 and 22.
%Third,
although we only eliminate 7 symmetries,
the best method offers a factor of 8 improvement
in search over not breaking symmetry.

To explore the interaction between symmetry breaking
and branching heuristics,
we report results in Table 4 to find the optimal
solution and prove optimality using different
branching heuristics.
We used the best two
symmetry breaking methods for the default left to right branching
heuristic (anti-gray inside-out, and anti-lex inside-out),
the worst symmetry breaking method for
the default branching heuristic (anti-gray outside-in),
a standard symmetry breaking method (lex),
the Gray code alternative (gray),
as well as no symmetry breaking (none).
We compared the default branching heuristic
(left2right heuristic) with branching heuristics
that instantiate variables right to left
(right2left heuristic),
alternating from both ends
inwards to the middle (outside-in heuristic),
from the middle alternating
outwards to both ends (inside-out heuristic),
by order of degree (degree heuristic),
and by order of the number of attached
constraints (constr heuristic).
Note that all domains are binary so there
is again no value for a heuristic like ff that
considers domain size.

%\myOmit
{
\begin{table}[htbp]
{\scriptsize
\begin{center}
\begin{tabular}{|c||r|r|r|r|r|r|} \hline
{Branching/SymBreak} & none & anti-gray & gray & lex & anti-gray & anti-lex \\                     &    & outside-in  &      &    & inside-out & inside-out \\
\hline
left2right heuristic & \ \ {\em 1,725,076} & \ \ 1,223,155 &   468,317 &   468,390 &   245,259 &   291,050 \\
right2left heuristic & {\em 1,725,076} &  {\em 322,291} &   329,897 &   299,877 & {\bf 224,540} &   269,628 \\
degree heuristic     & 2,024,484 &   603,857 &   329,897 &   400,228 &   500,415 & {\em 268,173} \\
constr heuristic     & 2,024,484 & 1,624,765 &   349,025 &   313,817 & \ \ 1,097,303 &   297,616 \\
inside-out heuristic & 1,786,741 & 2,787,164 & \ \ 1,406,831 & 1,055,918 &   326,938 &   268,206 \\
outside-in heuristic & 2,053,179 &   364,469 & {\em 284,417} &  {\em 284,526} & 2,044,042 & \ \ 2,767,059 \\
\hline
\end{tabular}
\end{center}
}
\caption{Backtracks required to find the 22 bit sequence
of lowest autocorrelation and prove optimality with different
branching heuristics and symmetry breaking constraints.
%Overall winner is in {\bf bold} font,
%whilst the column winners are in {\em emphasis}.
%Column max and min in { bold} font.
%``-'' indicates a time-out of 10 minutes.
}
\vspace{-3em}
\end{table}
}

\myOmit{
\begin{table}[htbp]
{\scriptsize
\begin{center}
\begin{tabular}{|c||r|r|r|r|r|r|} \hline
{Branching/SymBreak} & none & anti-gray & gray & lex & anti-gray & anti-lex \\
                     &    & outside-in  &      &    & inside-out & inside-out \\
\hline
inside-out heuristic & {\em 6,738,307} & 8,426,290 & 10,827,206 & 9,185,123 & 8,426,290 & 3,616,769 \\
left2right heuristic & 7,447,186 & 2,436,236 & 3,048,723 & 3,047,241 & 2,436,236 & 2,415,736 \\
right2left heuristic & 7,447,186 & 1,422,693 & 2,706,466 & 2,160,463 & 1,422,693 & 2,092,787 \\
outside-in heuristic & 9,945,023 & 1,312,986 &  {\em 2,286,197} & {\em 2,044,684} & 1,312,986 & 1,951,694 \\
degree heuristic     & 10,355,752 & 1,422,693 & 2,706,466 & 2,049,107 & 1,422,693 & 1,709,046 \\
constr heuristic     & 10,355,752 & {\em 1,242,244} & 2,632,703 & 2,297,652 & {\em 1,242,244} & {\bf 1,081,577} \\
\hline
\end{tabular}
\end{center}
}
\caption{Backtracks required to find the 24 bit sequence of lowest
autocorrelation and prove optimality with different branching heuristics and
symmetry breaking constraints.
%Overall winner is in {\bf bold} font,
%whilst the column winners are in {\em emphasis}.
%Column max and min in { bold} font.
%``-'' indicates a time-out of 10 minutes.
}
\vspace{-3em}
\end{table}
}

We make some observations about these results. First,
the best overall performance is observed when we break
symmetry with the anti-Gray code
ordering (anti-gray inside-out + right2left heuristic).
Second, we observe better performance
when the symmetry breaking constraint
aligns with the branching heuristic than
when it goes against it (e.g. anti-gray outside-in + outside-in heuristic is much better than anti-gray outside-in + inside-out heuristic).
%but lex inside-out + inside-out heuristic
%does much better than anti-lex inside-out + outside-in heuristic).
Third, the default heuristic (left2right) is
again competitive. %, but it is
%always beaten by the right2left heuristic.
%This
%is best or close to best with every symmetry breaking method.

\subsection{Peaceable armies of  queens}
\label{exp:queens}
The goal of this optimisation
problem is to place the largest
possible equal-sized armies of white and black
queens on a chess board so that no white queen
attacks a black queen or vice versa  \cite{bosch1}.
We used a simple model from
an earlier study of symmetry breaking
\cite{armies04}. The model has a
matrix of 0/1/2 decision variables,
in which $X_{ij}=2$ iff a black queen goes on square $(i,j)$,
$X_{ij}=1$ iff a white queen goes on square $(i,j)$,
and 0 otherwise.
Note that our model is now ternary, unlike the
binary models considered in the two previous
examples. However, the Gray code ordering extends from binary to ternary
codes in a straight forward. Similarly, we can
extend the decomposition to propagate Gray
code ordering constraints on ternary codes.

Table 5 reports results to find the optimal
solution and prove optimality for peaceable
armies of queens.
This model has 15 non-identity symmetries,
consisting of any combination of the
symmetries of the square and the
symmetry that swaps white queens for black queens.
We broke all 15 symmetries
by posting constraints to ensure that we only
find the smallest solution
in each symmetry class according to
the lexicographical or Gray code orderings (lex, gray),
or the largest solution in each
symmetry class according to the two orders (anti-lex, anti-gray).
We also considered symmetry breaking constraints
that take the variables in
row-wise order (row), in column-wise order (col),
in a snake order along the rows (snake),
in a snake order along the columns (col-snake),
or in a clockwise spiral (spiral).
%Finally, we considered
%symmetry breaking constraints that reversed the order of
%the variables (rev).
We again used the default variable
ordering that instantiates variables
in the lexicographical row-wise order.

\begin{table}[htbp]
{\scriptsize
\begin{center}
\begin{tabular}{|c||r|r|r|r|r|r|} \hline
{Symmetry breaking} & %3 &
$n=3$ & 4 & 5 & 6 & 7 & 8  \\ \hline \hline
none     & 19 & \ \ 194 & \ \ 2,588 & \ \ 37,434 & \ \ 679,771 & \ \ 19,597,858\\
lex col-snake & 13 & 98 & 1,014 & 8,638 & 199,964 & 5,299,787 \\
lex col & 23 &  87 & 1,042 & 10,792 & 198,032 & 5,197,013 \\
gray col & 26 & 101 & 1,118 & 9,763 & 214,391 & 5,008,279 \\
gray col-snake & 13 & 100 & 1,059 & 8,973 & 205,453 & 4,877,014 \\
gray spiral & 18 & 104 & 913 & 10,795 & 169,725 & 4,690,071 \\
lex spiral & 18 & 93 & 887 & 10,694 & 169,293 & 4,674,458 \\
gray row   & 19 &  73 &   680 & 6,975 & 116,725 & 3,705,591 \\
gray snake & 19 & 81 & 685 & 7,070 & 117,489 & 3,683,558\\
lex snake & 19 & 80 &  661 & 7,043 & 117,590 & 3,682,438 \\
lex row    & 19 &  73 &   679 & 6,880 & 115,999 & 3,652,269 \\
anti-gray spiral & 8 & 43 & 466 & 4,381 & 108,214 & 2,402,049 \\
anti-gray snake & 8 & 47 & 472 & 4,333 & 106,317 & 2,367,290 \\
anti-gray row & 8  &  44 &   452 & 4,326 & 105,837 & 2,357,024 \\
anti-lex col-snake & 18 & 59 & 560 & 4,513 & 70,950 & 2,346,875 \\
anti-lex col & 18 &  57 & 485 & 4,373 &  69,484 & 2,291,512 \\
anti-lex row & 9  &  29 &   315 & 3,417 & 101,530 & 2,037,336 \\
anti-lex snake & 9 & 34 & 314 & 3,366 & 100,472 & 2,010,354 \\
anti-lex spiral & 9 & 30 & 326 & 3,432 & 105,717 & 2,007,586 \\
anti-gray col-snake & 19 & 40 & 471 & 4,061 & 71,079 & 1,709,744 \\
anti-gray col & 19 & 40 & 385 & 4,317 &  70,632 & 1,698,492 \\
\hline
\end{tabular}
\end{center}
}
\caption{Backtracks required to solve the $n$ by $n$ peaceable
armies of queens problem to optimality with the default branching
heuristic.
%Column max and min in { bold} font.
%``-'' indicates a time-out of 10 minutes.
}
\vspace{-3em}
\end{table}

We make some observations about these results.
First, finding the largest solution in each symmetry
class (anti-gray and anti-lex) is always better
than finding the smallest (gray and lex).
We conjecture that this is because symmetry breaking lines up better
with the objective of maximizing the number of
queens on the board.
Second, symmetry breaking in
a ``conventional'' way (lex, row)
is beaten by half of the symmetry
breaking methods. In particular, all 10 methods which
find the largest solution up to symmetry in the Gray order (anti-gray)
or lexicographical ordering (anti-lex)
beat the ``conventional'' method (lex row).
Third, ordering the variables row-wise
in the symmetry breaking constraint is
best for lex, but for every other ordering (anti-lex, gray, anti-gray)
ordering variables row-wise
is never best. In particular,
anti-lex spiral beats anti-lex row and all other anti-lex methods,
gray snake beats gray row and all other gray methods,
and
anti-gray col beats anti-gray row and all other anti-gray methods.
Fourth, a good symmetry breaking method (e.g. anti-gray col)
offers up to a 12-fold improvement over not breaking the
15 non-identity symmetries.

To explore the interaction between symmetry breaking
and branching heuristics,
we report results in Table 6 using different
branching heuristics.
We used the best
symmetry breaking method for the default row-wise branching
heuristic (anti-gray col),
the worst symmetry breaking method for
the default branching heuristic (lex col-snake),
a standard method (lex row),
the Gray code alternative (gray row),
as well as no symmetry breaking (none).
We compared the same
branching heuristics as with the maximum density
still life problem. As domains are now
not necessarily binary, we also included
the ff heuristic that
order variables by their domain size tie-breaking
with the row heuristic (ff heuristic).
Given the good performance of the spiral and ff
heuristics individually, we also tried a novel heuristic
that combines them together,
branching on variables by order of the domain size
and tie-breaking with the spiral-in heuristic (ff-spiral heuristic).

\myOmit{
\begin{table}[htbp]
{\scriptsize
\begin{center}
\begin{tabular}{|c||r|r|r|r|r|} \hline
{Branching/SymBreak} & none & lex col-snake & gray row & lex row & anti-gray col \\ \hline
col-snake heuristic & 686,349 & {167,894} & 284,457 & 260,724 & 410,312 \\
col heuristic & 679,771 & {168,728} & 286,986 & 257,255 & 396,944 \\
spiral-out heuristic & 358,713 & {185,771} & 235,222 & 234,298 & 339,754 \\
degree heuristic & 679,771 & {158,145} & 177,060 & 177,203 & 266,158 \\
constr heuristic & {\em 315,225} & {105,871} & 111,999 & 111,130 & 115,002 \\
snake heuristic & 686,349 & 208,548 & 126,603 & 124,789 & {77,510} \\
row heuristic & 679,771 & 199,964 & 116,725 & 115,999 & {70,632} \\
ff heuristic & 569,811 & 157,495 & {\em 101,999} & {\em 104,290} & {65,527} \\
ff-spiral heuristic & 589,311 & 121,552 & 150,416 & 151,015 & 65,391 \\
spiral-in heuristic & 601,134 & {\em 89,263} & 121,465 & 121,160 & {\bf 59,188} \\
\hline
\end{tabular}
\end{center}
}
\caption{Backtracks required to solve the $7$ by $7$ peaceable
armies of queens problem to optimality for different branching
heuristics and symmetry breaking constraints.
%Overall winner is in {\bf bold} font,
%whilst the column winners are in {\em emphasis}.
%Column max and min in { bold} font.
%``-'' indicates a time-out of 10 minutes.
}
\vspace{-3em}
\end{table}
}

\begin{table}[htbp]
{\scriptsize
\begin{center}
\begin{tabular}{|c||r|r|r|r|r|} \hline
{Branching/SymBreak} & none & lex col-snake & gray row & lex row & anti-gray col \\ \hline
col-snake heuristic & \ \ 20,209,357 & 4,270,637 & \ \ 6,372,404 & \ \ 5,836,975 & 7,363,488 \\
col heuristic & 19,597,858 & 4,384,086 & 6,338,413 & 5,775,781 & 6,811,345\\
spiral-out heuristic & 8,196,693 & 4,894,264 & 5,099,899 & 5,126,074 & 6,478,506\\
degree heuristic & 19,597,858 & 3,129,599 & 4,216,463 & 4,343,792 & 6,351,547 \\
snake heuristic & 20,209,357 & 5,261,095 & 4,258,903 & 4,221,336 & 1,946,556 \\
constr heuristic & {\em 7,305,061} & 2,757,360 & 2,650,590 & 2,645,054 & 1,789,444 \\
row heuristic & 19,597,858 & 5,299,787 & 3,705,591 & 3,652,269 & 1,698,492 \\
ff heuristic & 12,826,856 & 3,371,419 & 2,495,788 & 2,521,351 & 1,309,529 \\
ff-spiral heuristic & 13,400,485 & 2,447,867 & 3,147,237 & {\em 2,162,657} & 1,222,607 \\
spiral-in heuristic & 15,577,982 & {\em 1,787,653} & {\em 2,387,067} & 2,430,499 & {\bf 1,193,988} \\
\hline
\end{tabular}
\end{center}
}
\caption{Backtracks required to solve the $8$ by $8$ peaceable
armies of queens problem to optimality for different branching
heuristics and symmetry breaking constraints.
%Overall winner is in {\bf bold} font,
%whilst the column winners are in {\em emphasis}.
%Column max and min in { bold} font.
%``-'' indicates a time-out of 10 minutes.
}
\vspace{-3em}
\end{table}

We make some observations about these results.
First, the best symmetry breaking constraint
with the default branching heuristic (anti-gray col + row heuristic)
was either very good or very bad with the other
branching heuristics. It offers the
best overall performance in this experiment
(viz. anti-gray col  + spiral-in heuristic), and is the best of all
the symmetry breaking methods for 5 other heuristics.
However, it also the worst of all the symmetry breaking
methods with 4 other heuristics.
Second, aligning the branching heuristic with the
symmetry breaking constraint at best offers
middle of the road performance (e.g. lex row + row heuristic)
but can also be counter-productive
(e.g. anti-gray col + col heuristic).
Third, the spiral-in heuristic
offer some of the best performance.
This heuristic provided the best
overall result, and was always in the top 2
for every symmetry breaking method.
Recall that the spiral-in heuristic
was one of the worst heuristics on the maximum density
still life problem.
We conjecture that this is because
it delays constraint propagation on the
still life problem constraints but not on the
constraints in the peaceable armies of queens problem.
Fourth, a bad combination of branching
heuristic and symmetry breaking constraints
is worse than
not breaking symmetry if we have a
good branching heuristic (e.g. none + constr heuristic
beats anti-gray col + col-snake heuristic).
}

These results support both
our hypotheses. Other orderings besides
the simple lexicographical ordering can be effective for
breaking symmetry, and symmetry breaking should
align with both the branching heuristic and the objective
function. Unfortunately, as
the last example demonstrated,
the interaction between problem constraints,
symmetry breaking and branching heuristic can be
complex and difficult to predict.
Overall, the Gray code ordering
appears useful. Whilst it is conceptually
similar to the lexicographical ordering,
it looks at more than one bit at a time. 
This is reflected in the automaton for 
the Gray code ordering which has more states
than that required for the lexicographical ordering.

\section{Conclusions}

We have argued that in general breaking
symmetry with a different ordering over assignments
than the usual lexicographical
ordering %used with the Lex-Leader method
does not improve the computational complexity of breaking with symmetry.
Our argument had two parts.
First, we argued that under modest assumptions
we cannot reduce the worst case complexity from that
of breaking symmetry with a lexicographical ordering.
These assumptions are satisfied by the Gray code and
Snake-Lex orderings. Second, we proved
that for the particular case of row and column
symmetries, breaking symmetry with the Gray code or Snake-Lex
ordering is intractable (as it was with the
lexicographical ordering).
We then explored algorithms to break
symmetry with other orderings.
In particular, we gave a linear time propagator for the Gray code
ordering constraint, and proved that enforcing domain consistency
on the $\snakelex$ constraint, like on the $\DLex$ constraint,
is NP-hard.
Finally, we demonstrated that
other orderings have promise in practice.
We ran experiments on three standard benchmark
domains where breaking symmetry with the Gray code ordering
was often better than with the Lex-Leader or Snake-Lex methods.
%An interesting application for breaking
%symmetry with other orderings would be
%to dynamic methods that select the
%ordering based on how search is
%progressing.

%\section*{Acknowledgements}

%\bibliographystyle{named}
\bibliographystyle{splncs}
%\bibliography{/Users/twalsh/Documents/biblio/a-z2,/Users/twalsh/Documents/biblio/pub2}
%\bibliography{a-z2,pub2}

\end{document}